\documentclass[10pt,twocolumn,letterpaper]{article}
\usepackage[pagenumbers]{cvpr}
\usepackage{tikz}
\usepackage{colortbl}
\usepackage{amsthm}

\definecolor{tablegood}{rgb}{0.80,1.00,0.80}
\definecolor{tablemid}{rgb}{1.00,1.00,0.80}
\definecolor{tablepoor}{rgb}{1.00,0.80,0.80}
\newcommand{\metriccell}[2]{\cellcolor{#1}#2}
\newcommand{\tableformat}{\renewcommand{\arraystretch}{1.12}\setlength{\aboverulesep}{1.8pt}\setlength{\belowrulesep}{1.8pt}\setlength{\heavyrulewidth}{0.8pt}\setlength{\lightrulewidth}{0.4pt}}
\usetikzlibrary{positioning,arrows.meta}
\usepackage{pgfplots}
\pgfplotsset{compat=1.18}

\definecolor{cvprblue}{rgb}{0.21,0.49,0.74}
\usepackage[pagebackref,breaklinks,colorlinks,allcolors=cvprblue]{hyperref}
\title{SPAR: Self-Referenced World-Model Rewards for Physical Alignment in Video Generation}
\author{Junyi Zhang$^{1}$\quad Qichao Wang$^{1}$\quad Wenzhe Qian$^{1}$\quad Yuanchuan Wang$^{1}$\quad Min Zhang$^{1}$\\
$^{1}$Zhejiang University\\
{\tt\small gavin.jy@zju.edu.cn}}

\makeatletter

\g@addto@macro\@maketitle{\vspace{-20pt}\begin{minipage}{\textwidth}
\centering
\setlength{\belowcaptionskip}{0.5em}
\includegraphics[width=\linewidth]{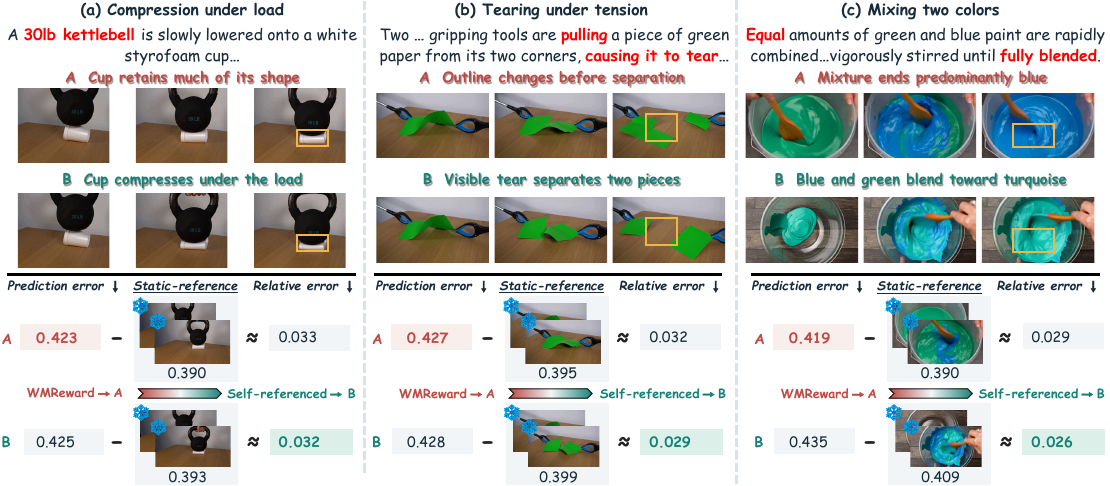}
\captionof{figure}{\textbf{Lower prediction error does not always imply more plausible dynamics.} WMReward favors candidate A in each example despite B showing more plausible physical evolution: the cup compresses under the load, the paper visibly tears apart, and the paints blend toward turquoise. Subtracting each candidate's own static-reference error reverses these preferences and favors B, illustrating how a content-matched reference changes the evaluation of generated dynamics. Yellow boxes highlight key visual differences.}
\label{fig:teaser}
\end{minipage}
\par
}
\makeatother
\begin{document}
\maketitle
\begin{abstract}
Generating videos with physically plausible dynamics remains challenging despite advances in visual quality.
Latent world models offer predictive signals for evaluating physical plausibility, but higher agreement between predicted and observed features may reflect easier-to-predict scene content rather than more plausible dynamics.
We propose \textbf{S}elf-Referenced \textbf{P}hysical \textbf{A}lignment \textbf{R}ewards (\textbf{SPAR}), a framework that uses world-model predictions for inference-time physical alignment.
SPAR compares each video's prediction error with that of static clips constructed by repeating its own frames, providing a content-matched reference for evaluating generated dynamics.
To account for the uneven spatial and temporal distribution of physical interactions, we introduce a unified \textbf{spatiotemporal reward refinement} strategy.
\textbf{Spatial refinement} emphasizes localized prediction discrepancies, while \textbf{temporal refinement} moderates brief error peaks and supplements prediction errors with agreement between predicted and observed feature changes.
Evaluations across multiple benchmarks demonstrate that SPAR improves the physical plausibility of both \textbf{text-to-video generation} and \textbf{image-to-video continuation}, bringing generated dynamics into closer alignment with the physical regularities governing real-world events.
Additional results and resources are available at \url{https://gavin-jy.github.io/spar-page/}.
\end{abstract}

\section{Introduction}
\label{sec:intro}

Physically realistic video generation is fundamental to using video models in embodied intelligence, robotics, and interactive simulation.
These applications require generated futures to capture how objects respond to forces, interact with their surroundings, and change state over time.
Despite substantial advances in visual quality, generated motion, contact, and deformation can still violate basic physical expectations~\cite{meng2024towards,motamed2026generative,bansal2026videophy}.
Ensuring that generated events follow plausible physical dynamics therefore remains a central challenge in video generation.

Existing approaches improve physical plausibility through generator training or inference-time alignment.
Training-based methods incorporate physical supervision and dynamics priors~\cite{li2025pisa,narayanan2026phyco,lu2026physis,lin2026mmphysvideo,yuan2026newtongen,zhu2026motionrag}, or optimize generators with rewards and preferences~\cite{wang2025physcorr,wang2026diffusion}.
These routes require additional optimization, often coupled with dedicated data curation, annotation, or simulation pipelines.
Inference-time alignment offers a complementary route, using reward signals to guide generation or select outputs~\cite{singhal2025general,yuan2026inference}.
A central challenge is designing rewards that assess the physical plausibility of generated dynamics.

Latent world models offer a promising basis for such rewards by learning to predict future representations from preceding observations~\cite{assran2025v}.
Their sensitivity to violations of object permanence and shape consistency~\cite{garrido2025intuitive} supports prediction--observation disagreement as evidence of departures from learned physical regularities.
WMReward~\cite{yuan2026inference} measures the cosine similarity between predicted and observed future representations, using their agreement as a reward for physical alignment.
However, predictive agreement depends on both scene content and temporal evolution.
Even under the same input condition, candidates can differ in appearance and scene configuration, so higher agreement may reflect easier-to-predict content rather than more plausible dynamics.
\textit{A physical reward should therefore account for content-dependent predictability when comparing generated dynamics.}

\textbf{Our key idea is to use static clips constructed from each candidate's own frames as a content-matched reference for evaluating its evolution.}
We introduce \textbf{S}elf-Referenced \textbf{P}hysical \textbf{A}lignment \textbf{R}ewards (\textbf{SPAR}) to implement this principle for inference-time physical alignment.
SPAR constructs static reference clips by repeating selected frames from each candidate, preserving their appearance while holding the input constant over time.
SPAR compares the world model's prediction errors on the original video and its static references, using the difference to form a self-referenced reward.
Figure~\ref{fig:teaser} illustrates how this self-referenced comparison favors plausible physical evolution over high absolute predictive agreement.

\begin{figure*}[t]
\centering
\includegraphics[width=\textwidth]{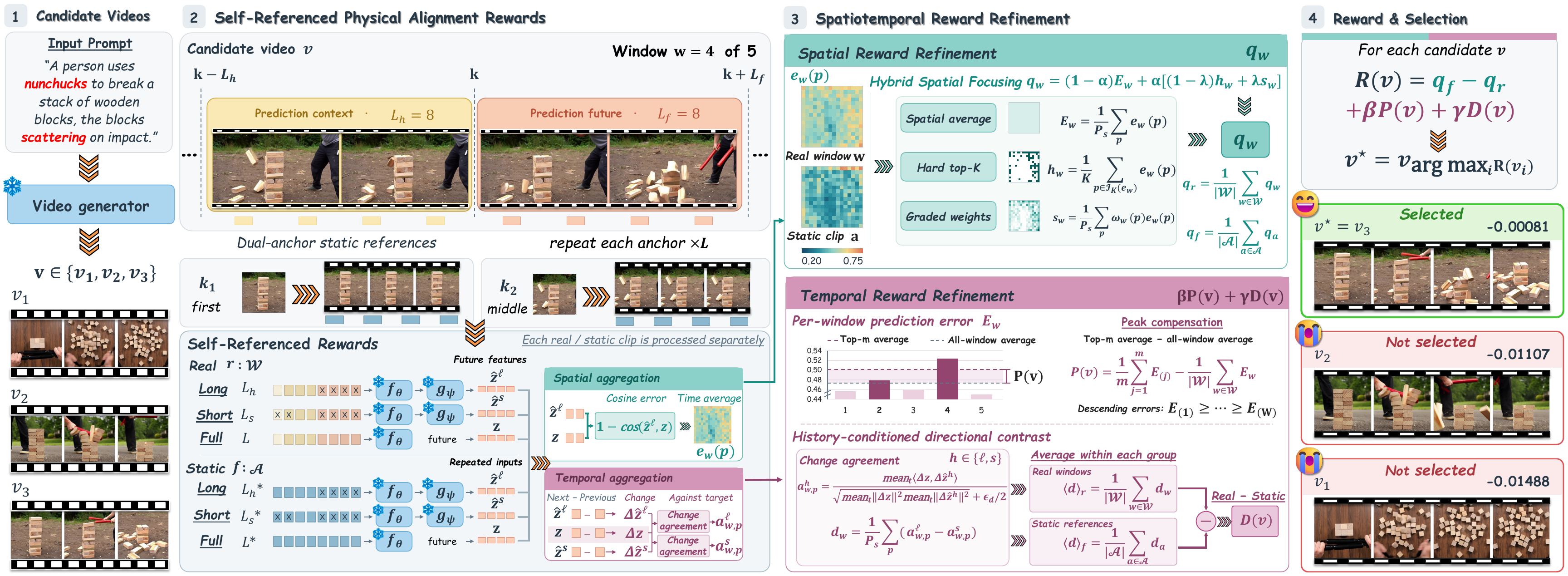}
\caption{\textbf{Overview of SPAR.} (1) Generate candidate videos from the same input. (2) Construct static references from each candidate's first and middle frames and evaluate both branches with a shared world model. (3) Refine spatial and temporal evidence through spatial focusing, peak compensation, and history-conditioned directional contrast. (4) Combine the reference-relative spatial and history terms with temporal peak compensation to score candidates and select the highest-reward video.}
\label{fig:pipeline}
\end{figure*}

The self-referenced reward provides a baseline for evaluating generated dynamics.
For physical alignment, its computation should also reflect the spatial and temporal structure of physical interactions.
Intuitively, these interactions are not uniformly distributed throughout a video: \textit{contact and deformation may be confined to small regions, while changes in motion or visibility may occur over brief intervals.}
This observation motivates our \textbf{spatiotemporal reward refinement} strategy for incorporating spatial discrepancies and temporal evolution into the self-referenced reward.
Spatial refinement combines the overall prediction error with greater emphasis on high-error regions, using both hard selection and graded weighting to preserve localized discrepancies.
The same computation is applied to the static references, allowing local evidence to enter the self-referenced comparison.
Temporal refinement moderates the contribution of high-error intervals to the mean prediction error, since brief changes in motion or visibility can challenge prediction even in plausible events.
It also complements error magnitudes with agreement between predicted and observed feature changes.
We compare this agreement under short and long histories, using the candidate's static references to calibrate the effect of earlier context.

We evaluate SPAR across multiple benchmarks and video generation tasks, comparing with both video generation models and reward-based alignment methods.
For text-to-video generation, SPAR achieves competitive scores of 65.00 on PhyGenBench~\cite{meng2024towards} and 62.78\% on VideoPhy-2 Hard~\cite{bansal2026videophy}.
We also evaluate image-to-video continuation on Physics-IQ Verified~\cite{radsch2026physics}, where SPAR achieves the highest overall score of 44.47 and leads all four continuation-similarity components among the evaluated methods.

The main contributions of our work are as follows:
\begin{itemize}
    \item We propose SPAR, a framework that uses self-referenced world-model rewards to improve the physical plausibility of generated videos through inference-time alignment.
    \item We develop a unified spatiotemporal reward refinement strategy that emphasizes local spatial discrepancies and captures temporal changes for physical alignment.
    \item We evaluate SPAR across multiple benchmarks and video generation tasks, demonstrating its effectiveness.
\end{itemize}

\section{Related Work}
\label{sec:related}

\paragraph{Physics-aware video generation.}
Physics-aware generation incorporates physical knowledge through supervision, representations, and optimization objectives.
Simulation provides controlled physical events for post-training in PISA~\cite{li2025pisa} and PhyCo~\cite{narayanan2026phyco}, while Physis-Lang~\cite{lu2026physis} connects physical descriptions with data retrieval, training, and inference-time prompting.
Representation-based approaches equip generators with perceptual cues~\cite{lin2026mmphysvideo}, learned dynamical constraints~\cite{yuan2026newtongen}, or adapted motion priors from retrieved videos~\cite{zhu2026motionrag}.
Physical evaluation also supplies training objectives: PhysCorr~\cite{wang2025physcorr} uses physical preferences, and Diffusion-DRF~\cite{wang2026diffusion} updates the generator with fine-grained feedback from a frozen VLM.
Training-based methods require additional optimization of the generator or auxiliary modules, often supported by specialized data or physical annotations.
SPAR instead derives physical alignment rewards from an existing world model and each candidate's own static references, without additional training.

\paragraph{Inference-time alignment reward.}
Inference-time alignment changes generation conditions, sampling, or output selection while keeping generator weights fixed.
Prompt refinement uses reasoning or video feedback~\cite{xue2025phyt2v,liu2025bootstrapping}, including separately trained refiners~\cite{wu2026phyprompt}, while sampling guidance uses descriptions of implausible behavior~\cite{hao2025enhancing} or planned keyframes and trajectories~\cite{zhuang2026causalmotion}.
Reward-guided approaches include Feynman--Kac steering~\cite{singhal2025general} for diffusion sampling and Seeking Physics in Diffusion Noise~\cite{tang2026seeking}, which trains a verifier on intermediate denoising features.
For evaluating generated videos, GeoPhys~\cite{interno2026geophys} derives rewards from the geometry of image-encoder feature trajectories, including speed variation, curvature, and linear prediction residuals.
WMReward~\cite{yuan2026inference} instead uses agreement between predicted and observed future representations from V-JEPA~2~\cite{assran2025v}.
However, absolute predictive agreement reflects both scene content and temporal evolution, so candidates with easier-to-predict content may receive higher rewards without exhibiting more plausible dynamics.
SPAR builds on predictive world-model signals by evaluating each candidate relative to its own static references and refining the spatial and temporal contributions to the reward.

\paragraph{Latent world models and physical understanding.}
Latent world models learn temporal regularities by predicting future representations from observed context.
V-JEPA~2~\cite{assran2025v} provides a learned representation space and a predictor for modeling such evolution.
Physical understanding studies examine whether predictive models respond to violations of expected dynamics.
IntPhys~2~\cite{bordes2025intphys} provides controlled possible and impossible events testing permanence, immutability, spatiotemporal continuity, and solidity.
Garrido \etal~\cite{garrido2025intuitive} show that self-supervised latent prediction can yield sensitivity to violations of object permanence and shape consistency.
These findings motivate predictive representations as a source of physical evaluation signals.
Our work connects this predictive physical understanding to the evaluation and selection of generated videos.
Appendix~\ref{sec:background} expands the background summarized here.

\section{Method}
\label{sec:method}

SPAR performs inference-time physical alignment of video generation through self-referenced world-model rewards.
As illustrated in Fig.~\ref{fig:pipeline}, the framework evaluates each candidate's evolution relative to its own static response and integrates spatial and temporal predictive evidence into the selection reward.
We first introduce best-of-$N$ selection and the world-model prediction errors used for scoring (Sec.~\ref{sec:preliminaries}).
We then establish the candidate-specific reference construction and relative reward (Sec.~\ref{sec:srpa}).
Building on this framework, we develop a unified spatiotemporal reward refinement strategy that organizes predictive evidence into the final alignment reward (Sec.~\ref{sec:aggregation}).

\subsection{Preliminaries: World-Model Rewards}
\label{sec:preliminaries}
\label{sec:features}

Best-of-$N$ (BoN) selection returns the highest-reward video among $N$ candidates $\{v_i\}_{i=1}^{N}$ generated under the same input condition.
WMReward~\cite{yuan2026inference} applies this strategy using the mean cosine similarity between predicted and observed future features as the reward.

Let $f_\theta$ and $g_\psi$ denote the frozen video encoder and latent predictor of V-JEPA~2~\cite{assran2025v}, respectively.
A video is sampled to $F$ frames at resolution $r\times r$ and partitioned into $W$ overlapping windows of length $L$, with stride $s$.
Within each window, the first $L_h$ frames provide the observed history and the remaining $L_f=L-L_h$ frames form the prediction target.
The encoder processes only the observed history tokens, and the predictor estimates features at the future temporal positions.
Target features are obtained by encoding the full window and retaining its future positions.
For temporal tubelet size $b$, the target contains $T_f=L_f/b$ temporal tokens at each of $P_s$ spatial positions.

Let $\hat z^{\ell}_{w,t,p}$ and $z_{w,t,p}$ denote predicted and target features after LayerNorm and $\ell_2$ normalization, where $w$ indexes windows, $t$ future temporal tokens, $p$ spatial positions, and $\ell$ the full observed history.
The spatial error map and its window mean are
\begin{equation}
\begin{aligned}
e_w(p)&=\frac1{T_f}\sum_{t=1}^{T_f}
\left(1-\left\langle\hat z^{\ell}_{w,t,p},z_{w,t,p}\right\rangle\right),\\
E_w&=\frac1{P_s}\sum_p e_w(p).
\end{aligned}
\label{eq:error}
\end{equation}
The WMReward baseline aggregates these errors uniformly over the $W$ windows:
\begin{equation}
R_{\mathrm{abs}}(v)=-\frac1W\sum_{w=1}^{W}E_w
=-\frac{1}{WP_s}\sum_{w,p}e_w(p).
\label{eq:absolute}
\end{equation}
This score equals the mean cosine similarity minus one and therefore preserves the same candidate ranking.

For best-of-$N$ physical alignment, prediction errors must support comparisons between candidates whose scene content and temporal evolution can both vary, even under the same input condition.
Content-dependent prediction difficulty can therefore differ across candidates: a lower absolute error may reflect easier-to-predict content as well as more plausible dynamics.
This motivates interpreting each candidate's prediction error relative to a baseline matched to its own scene content.

\subsection{Self-Referenced Physical Alignment Rewards}
\label{sec:calibration}
\label{sec:srpa}

We construct a content-matched baseline by repeating selected frames from each candidate video to form static reference clips.
These clips preserve the appearance of their source frames while holding the input constant over time.
Applying the same frozen world model and predictive readout to the original video and its static references defines the reward as a difference between their prediction errors.

\vspace{-1em}
\paragraph{Dual-anchor reference construction.}
We select the first and middle frames of the sampled video as anchor frames, indexed by $a_1$ and $a_2$.
Each anchor is repeated $L$ times and processed with the same encoder, predictor, and history mask as a real window.

\vspace{-1em}
\paragraph{Reference-relative scoring.}
Let $\mathcal W$ contain the $W$ real windows and $\mathcal A$ the two static clips.
For any window statistic $\xi$, define the equal-weight averages
\begin{equation}
\langle \xi\rangle_{\mathrm r}=\frac1{|\mathcal W|}\sum_{w\in\mathcal W}\xi_w,
\qquad
\langle \xi\rangle_{\mathrm f}=\frac1{|\mathcal A|}\sum_{a\in\mathcal A}\xi_a.
\end{equation}
Using mean prediction error as the readout, we define the self-referenced reward as
\begin{equation}
B(v)=\langle E\rangle_{\mathrm f}-\langle E\rangle_{\mathrm r}.
\label{eq:baseline}
\end{equation}
A higher reward indicates better prediction of the evolving video relative to its own static references.

A shared reference would add the same offset to every candidate's score and leave their ranking unchanged.
Candidate-specific references instead adjust the comparison according to each video's own static prediction error.
We next extend this self-referenced comparison to spatial and temporal predictive evidence.

\subsection{Spatiotemporal Reward Refinement}
\label{sec:aggregation}

WMReward computes its reward by uniformly averaging prediction errors across spatial positions and temporal windows.
The self-referenced reward in Sec.~\ref{sec:srpa} introduces a candidate-specific baseline while retaining this averaging scheme.
Such averaging does not explicitly distinguish the spatial distribution and temporal structure of predictive discrepancies.
We refine the reward along spatial and temporal dimensions: spatial refinement emphasizes localized discrepancies, while temporal refinement accounts for error concentration and history-conditioned feature changes.
\vspace{-1em}
\paragraph{Spatial Reward Refinement.}
Uniform spatial averaging can dilute localized discrepancies among many low-error positions.
We therefore assign greater weight to high-error positions while retaining the global mean, so that both overall prediction error and local discrepancies contribute to the score.
Our Hybrid Spatial Focusing formulates this allocation as a fixed spatial weight budget under a per-position cap.
Within this formulation, hard selection concentrates weight on the largest errors, whereas a quadratic regularizer yields graded weights based on their relative magnitudes.

Let $\mathcal I_K(e_w)$ index the $K$ largest errors among the $P_s$ positions.
Assigning equal weight to these positions gives
\begin{equation}
h_w=\frac1K\sum_{p\in\mathcal I_K(e_w)}e_w(p).
\end{equation}

For the regularized allocation, standardize the errors as $u_w(p)=(e_w(p)-E_w)/\max(\sigma_w,\epsilon_s)$, where $\sigma_w$ is the population standard deviation and $\epsilon_s>0$ stabilizes division.
The resulting weights and weighted mean are
\begin{equation}
\begin{aligned}
\omega_w(p)&=\min\{c,\max\{0,u_w(p)-\tau_w\}\},\\
\frac1{P_s}\sum_p\omega_w(p)&=1,\qquad
s_w=\frac1{P_s}\sum_p\omega_w(p)e_w(p).
\end{aligned}
\end{equation}
Here $c\geq1$ caps each weight and $\tau_w$ enforces unit mean weight.
When $c=P_s/K$, as in our default configuration, hard selection and continuous weighting solve the unregularized and quadratically regularized allocation problems under the same constraints.

Both allocations produce normalized weighted averages.
We interpolate between their concentration patterns and combine the result with the global mean:
\begin{equation}
q_w=(1-\alpha)E_w+\alpha\big[(1-\lambda)h_w+\lambda s_w\big].
\label{eq:windowreadout}
\end{equation}
Here $\alpha$ balances global and local error contributions, while $\lambda$ interpolates between hard and regularized weight allocations.
Each real or static window computes its own weights, so the comparison operates on error distributions without requiring correspondence between moving objects and anchor positions.
The constrained optimization formulation is provided in the supplementary material (Eq.~\ref{eq:softobjective}).
\vspace{-1em}
\paragraph{Temporal Reward Refinement.}
Temporal refinement considers both how prediction errors are distributed across windows and how predicted feature changes agree with the observed evolution.

Brief changes in motion or visibility can be difficult to predict even in plausible events, allowing a few high-error windows to dominate the video-level assessment.
Temporal Peak Compensation moderates the contribution of high-error windows using the concentration statistic
\begin{equation}
P(v)=\operatorname{TopMean}_m\{E_w:w\in\mathcal W\}
-\langle E\rangle_{\mathrm r},
\label{eq:peak}
\end{equation}
where $\operatorname{TopMean}_m$ averages the $m$ largest real-window errors, with $1\leq m<W$.
Adding $\beta P(v)$ redistributes the mean-error penalty away from the highest-error windows while preserving the additional spatial contribution $q_w-E_w$.
The equivalent weights are derived from Eq.~\ref{eq:temporal-reweighting} in the supplementary material.

Error magnitudes do not explicitly measure whether predicted feature changes align with observed changes.
We measure this alignment through temporal feature differences, then compare long- and short-history predictions to quantify the contribution of earlier context to directional agreement.
Both predictions target the same future positions and share the target features, with the short history retaining the last $L_s<L_h$ observed frames.

For history $h\in\{\ell,s\}$, let $\Delta z$ and $\Delta\hat z^h$ be first differences across the $T_f$ future tokens.
Their normalized agreement at each spatial position is
\begin{equation}
a^h_{w,p}=
\frac{\operatorname{mean}_t\langle\Delta z_{w,t,p},\Delta\hat z^h_{w,t,p}\rangle}
{\sqrt{\operatorname{mean}_t\|\Delta z_{w,t,p}\|^2
\operatorname{mean}_t\|\Delta\hat z^h_{w,t,p}\|^2}+\epsilon_d/2},
\end{equation}
where $\epsilon_d>0$ stabilizes normalization.
The model also exhibits history-conditioned feature-change agreement on repeated-frame inputs, so a long-history gain can occur without input motion.
We define History-Conditioned Directional Contrast by comparing the two histories and using the static clips as a candidate-specific baseline:
\begin{equation}
d_w=\frac1{P_s}\sum_p(a^\ell_{w,p}-a^s_{w,p}),
\qquad
D(v)=\langle d\rangle_{\mathrm r}-\langle d\rangle_{\mathrm f}.
\end{equation}
Thus, $D(v)$ measures the context response of feature-change agreement relative to the candidate's static baseline.
\vspace{-1em}
\paragraph{Final Reward and Selection.}
The complete reward combines reference-relative error readout, within-video concentration compensation, and calibrated history contrast:
\begin{equation}
R(v)=\langle q\rangle_{\mathrm f}-\langle q\rangle_{\mathrm r}
+\beta P(v)+\gamma D(v).
\label{eq:reward}
\end{equation}
Thus, spatial prediction errors and history-dependent feature changes share one calibration principle, while temporal concentration adjusts the mean-error contribution.
The coefficient $\beta$ controls concentration compensation, and $\gamma$ determines the strength and sign of the history-contrast contribution.
With the generator fixed, SPAR selects
\begin{equation}
v^\star=v_{\arg\max_i R(v_i)}.
\end{equation}

\section{Experiments}
\label{sec:experiments}

\begin{table*}[t]
\centering
\footnotesize
\tableformat
\setlength{\tabcolsep}{0.6pt}
\begin{minipage}[t]{0.49\textwidth}
\caption{Overall results. Each entry reports the \textbf{benchmark score $\uparrow$ / SPAR win rate $\uparrow$}. Colors rank each column from {\setlength{\fboxsep}{1pt}\colorbox{tablegood}{best}} to {\setlength{\fboxsep}{1pt}\colorbox{tablepoor}{worst}}.}
\label{tab:main}
\centering
\newcommand{\scorewin}[2]{\makebox[2.5em][c]{#1}\,/\,\makebox[2.2em][c]{#2}}
\begin{tabular*}{\linewidth}{@{}>{\raggedright\arraybackslash}p{\dimexpr.33\linewidth-\tabcolsep\relax}>{\centering\arraybackslash}p{\dimexpr.223\linewidth-2\tabcolsep\relax}>{\centering\arraybackslash}p{\dimexpr.223\linewidth-2\tabcolsep\relax}>{\centering\arraybackslash}p{\dimexpr.224\linewidth-2\tabcolsep\relax}@{\hspace{\tabcolsep}}}
\toprule
{\bfseries\boldmath Method} & {\bfseries\boldmath Physics-IQ} & {\bfseries\boldmath PhyGenBench} & {\bfseries\boldmath VideoPhy-2}\\
\midrule
\mbox{Wan2.2-TI2V-5B~\cite{wan2025wan}} & \metriccell{tablepoor!71!tablemid}{\scorewin{30.79}{78\%}} & \metriccell{tablegood!33!tablemid}{\scorewin{63.75}{54\%}} & \metriccell{tablepoor!43!tablemid}{\scorewin{27.22}{90\%}}\\
\mbox{CogVideoX1.5-5B~\cite{yang2025cogvideox}} & \metriccell{tablepoor!100!tablemid}{\scorewin{17.97}{91\%}} & \metriccell{tablepoor!67!tablemid}{\scorewin{60.83}{60\%}} & \metriccell{tablepoor!71!tablemid}{\scorewin{22.78}{91\%}}\\
\mbox{Cosmos3-Nano~\cite{agarwal2026cosmos}} & \metriccell{tablepoor!43!tablemid}{\scorewin{33.07}{72\%}} & \metriccell{tablepoor!100!tablemid}{\scorewin{56.46}{68\%}} & \metriccell{tablepoor!100!tablemid}{\scorewin{16.11}{94\%}}\\
\mbox{HunyuanVideo-1.5~\cite{wu2025hunyuanvideo}} & \metriccell{tablepoor!14!tablemid}{\scorewin{34.83}{72\%}} & \metriccell{tablegood!100!tablemid}{\scorewin{\textbf{65.83}}{48\%}} & \metriccell{tablepoor!14!tablemid}{\scorewin{53.33}{62\%}}\\
\mbox{MiniMax-H3~\cite{minimaxh3}} & \metriccell{tablegood!14!tablemid}{\scorewin{40.59}{66\%}} & \metriccell{tablegood!0!tablemid}{\scorewin{62.08}{66\%}} & \metriccell{tablegood!43!tablemid}{\scorewin{59.81}{57\%}}\\
\mbox{GeoPhys~\cite{interno2026geophys}} & \metriccell{tablegood!43!tablemid}{\scorewin{42.62}{61\%}} & \metriccell{tablegood!33!tablemid}{\scorewin{63.75}{58\%}} & \metriccell{tablegood!14!tablemid}{\scorewin{58.33}{61\%}}\\
\mbox{WMReward~\cite{yuan2026inference}} & \metriccell{tablegood!71!tablemid}{\scorewin{\underline{42.88}}{64\%}} & \metriccell{tablepoor!33!tablemid}{\scorewin{61.88}{73\%}} & \metriccell{tablegood!71!tablemid}{\scorewin{\underline{61.11}}{56\%}}\\
\mbox{SPAR (ours)} & \metriccell{tablegood!100!tablemid}{\scorewin{\textbf{44.47}}{\textemdash}} & \metriccell{tablegood!67!tablemid}{\scorewin{\underline{65.00}}{\textemdash}} & \metriccell{tablegood!100!tablemid}{\scorewin{\textbf{62.78}}{\textemdash}}\\
\bottomrule
\end{tabular*}
\end{minipage}\hfill
\begin{minipage}[t]{0.49\textwidth}
\caption{PhyGenBench results. Scores cover five physical categories. Chem. and Phys. denote chemical and physical properties.}
\label{tab:pgdetail}
\centering
\begin{tabular*}{\linewidth}{@{}>{\raggedright\arraybackslash}p{\dimexpr0.33\linewidth-\tabcolsep\relax}*{5}{>{\raggedleft\arraybackslash}p{\dimexpr0.109\linewidth-2\tabcolsep\relax}}>{\raggedleft\arraybackslash}p{\dimexpr0.125\linewidth-2\tabcolsep\relax}@{\hspace{\tabcolsep}}}
\toprule
{\bfseries\boldmath Method} & {\bfseries\boldmath Chem.} & {\bfseries\boldmath Force} & {\bfseries\boldmath Heat} & {\bfseries\boldmath Light} & {\bfseries\boldmath Phys.} & {\bfseries\boldmath Overall}\\
\midrule
\mbox{Wan2.2-TI2V-5B~\cite{wan2025wan}} & \metriccell{tablegood!60!tablemid}{60.00} & \metriccell{tablegood!71!tablemid}{65.00} & \metriccell{tablegood!67!tablemid}{61.11} & \metriccell{tablepoor!20!tablemid}{70.00} & \metriccell{tablegood!0!tablemid}{54.67} & \metriccell{tablegood!33!tablemid}{63.75}\\
\mbox{CogVideoX1.5-5B~\cite{yang2025cogvideox}} & \metriccell{tablegood!60!tablemid}{60.00} & \metriccell{tablepoor!100!tablemid}{57.50} & \metriccell{tablegood!100!tablemid}{64.44} & \metriccell{tablepoor!100!tablemid}{66.00} & \metriccell{tablepoor!67!tablemid}{52.00} & \metriccell{tablepoor!67!tablemid}{60.83}\\
\mbox{Cosmos3-Nano~\cite{agarwal2026cosmos}} & \metriccell{tablepoor!100!tablemid}{42.22} & \metriccell{tablepoor!43!tablemid}{59.17} & \metriccell{tablepoor!100!tablemid}{50.00} & \metriccell{tablepoor!60!tablemid}{66.67} & \metriccell{tablepoor!100!tablemid}{48.00} & \metriccell{tablepoor!100!tablemid}{56.46}\\
\mbox{HunyuanVideo-1.5~\cite{wu2025hunyuanvideo}} & \metriccell{tablepoor!20!tablemid}{53.33} & \metriccell{tablegood!100!tablemid}{65.83} & \metriccell{tablegood!0!tablemid}{57.78} & \metriccell{tablegood!100!tablemid}{79.33} & \metriccell{tablegood!33!tablemid}{56.00} & \metriccell{tablegood!100!tablemid}{\textbf{65.83}}\\
\mbox{MiniMax-H3~\cite{minimaxh3}} & \metriccell{tablepoor!60!tablemid}{52.59} & \metriccell{tablepoor!71!tablemid}{58.89} & \metriccell{tablepoor!33!tablemid}{56.67} & \metriccell{tablegood!20!tablemid}{74.67} & \metriccell{tablepoor!33!tablemid}{54.22} & \metriccell{tablegood!0!tablemid}{62.08}\\
\mbox{GeoPhys~\cite{interno2026geophys}} & \metriccell{tablegood!100!tablemid}{62.22} & \metriccell{tablegood!14!tablemid}{61.67} & \metriccell{tablepoor!67!tablemid}{54.44} & \metriccell{tablegood!60!tablemid}{76.00} & \metriccell{tablegood!0!tablemid}{54.67} & \metriccell{tablegood!33!tablemid}{63.75}\\
\mbox{WMReward~\cite{yuan2026inference}} & \metriccell{tablegood!20!tablemid}{55.56} & \metriccell{tablepoor!14!tablemid}{60.00} & \metriccell{tablepoor!100!tablemid}{50.00} & \metriccell{tablegood!20!tablemid}{74.67} & \metriccell{tablegood!67!tablemid}{57.33} & \metriccell{tablepoor!33!tablemid}{61.88}\\
\mbox{SPAR (ours)} & \metriccell{tablegood!60!tablemid}{60.00} & \metriccell{tablegood!43!tablemid}{63.33} & \metriccell{tablegood!33!tablemid}{58.89} & \metriccell{tablegood!20!tablemid}{74.67} & \metriccell{tablegood!100!tablemid}{58.67} & \metriccell{tablegood!67!tablemid}{\underline{65.00}}\\
\bottomrule
\end{tabular*}
\end{minipage}
\par
\vspace{8pt}
\begin{minipage}[t]{0.49\textwidth}
\caption{VideoPhy-2 Hard. SA: semantic adherence. PC: physical commonsense. Overall counts videos with both scores at least four.}
\label{tab:vpdetail}
\centering
\begin{tabular*}{\linewidth}{@{}>{\raggedright\arraybackslash}p{\dimexpr0.36\linewidth-\tabcolsep\relax}*{5}{>{\raggedleft\arraybackslash}p{\dimexpr0.1280000\linewidth-2\tabcolsep\relax}}@{\hspace{\tabcolsep}}}
\toprule
{\bfseries\boldmath Method} & {\bfseries\boldmath SA} & {\bfseries\boldmath PC} & {\bfseries\boldmath SA$\geq4$} & {\bfseries\boldmath PC$\geq4$} & {\bfseries\boldmath Overall}\\
\midrule
\mbox{Wan2.2-TI2V-5B~\cite{wan2025wan}} & \metriccell{tablepoor!43!tablemid}{3.02} & \metriccell{tablepoor!71!tablemid}{4.03} & \metriccell{tablepoor!43!tablemid}{35.56} & \metriccell{tablepoor!67!tablemid}{73.33} & \metriccell{tablepoor!43!tablemid}{27.22}\\
\mbox{CogVideoX1.5-5B~\cite{yang2025cogvideox}} & \metriccell{tablepoor!71!tablemid}{2.92} & \metriccell{tablepoor!100!tablemid}{3.90} & \metriccell{tablepoor!71!tablemid}{31.11} & \metriccell{tablepoor!100!tablemid}{68.33} & \metriccell{tablepoor!71!tablemid}{22.78}\\
\mbox{Cosmos3-Nano~\cite{agarwal2026cosmos}} & \metriccell{tablepoor!100!tablemid}{2.49} & \metriccell{tablegood!14!tablemid}{4.42} & \metriccell{tablepoor!100!tablemid}{18.89} & \metriccell{tablepoor!33!tablemid}{83.33} & \metriccell{tablepoor!100!tablemid}{16.11}\\
\mbox{HunyuanVideo-1.5~\cite{wu2025hunyuanvideo}} & \metriccell{tablepoor!14!tablemid}{3.68} & \metriccell{tablepoor!14!tablemid}{4.41} & \metriccell{tablepoor!14!tablemid}{60.56} & \metriccell{tablegood!0!tablemid}{83.89} & \metriccell{tablepoor!14!tablemid}{53.33}\\
\mbox{MiniMax-H3~\cite{minimaxh3}} & \metriccell{tablegood!71!tablemid}{4.04} & \metriccell{tablegood!43!tablemid}{4.54} & \metriccell{tablegood!14!tablemid}{70.37} & \metriccell{tablegood!33!tablemid}{86.48} & \metriccell{tablegood!43!tablemid}{59.81}\\
\mbox{GeoPhys~\cite{interno2026geophys}} & \metriccell{tablegood!100!tablemid}{4.05} & \metriccell{tablepoor!43!tablemid}{4.39} & \metriccell{tablegood!43!tablemid}{70.56} & \metriccell{tablepoor!33!tablemid}{83.33} & \metriccell{tablegood!14!tablemid}{58.33}\\
\mbox{WMReward~\cite{yuan2026inference}} & \metriccell{tablegood!43!tablemid}{4.02} & \metriccell{tablegood!71!tablemid}{4.54} & \metriccell{tablegood!100!tablemid}{71.67} & \metriccell{tablegood!67!tablemid}{86.67} & \metriccell{tablegood!71!tablemid}{\underline{61.11}}\\
\mbox{SPAR (ours)} & \metriccell{tablegood!14!tablemid}{3.98} & \metriccell{tablegood!100!tablemid}{4.62} & \metriccell{tablegood!71!tablemid}{71.11} & \metriccell{tablegood!100!tablemid}{88.89} & \metriccell{tablegood!100!tablemid}{\textbf{62.78}}\\
\bottomrule
\end{tabular*}
\end{minipage}\hfill
\begin{minipage}[t]{0.49\textwidth}
\caption{Physics-IQ Verified results. ST, SP, WS, and MSE measure continuation similarity. Overall averages the four components.}
\label{tab:iqdetail}
\centering
\begin{tabular*}{\linewidth}{@{}>{\raggedright\arraybackslash}p{\dimexpr0.36\linewidth-\tabcolsep\relax}*{5}{>{\raggedleft\arraybackslash}p{\dimexpr0.1280000\linewidth-2\tabcolsep\relax}}@{\hspace{\tabcolsep}}}
\toprule
{\bfseries\boldmath Method} & {\bfseries\boldmath ST} & {\bfseries\boldmath SP} & {\bfseries\boldmath WS} & {\bfseries\boldmath MSE} & {\bfseries\boldmath Overall}\\
\midrule
\mbox{Wan2.2-TI2V-5B~\cite{wan2025wan}} & \metriccell{tablepoor!14!tablemid}{28.11} & \metriccell{tablepoor!71!tablemid}{39.06} & \metriccell{tablepoor!71!tablemid}{27.38} & \metriccell{tablepoor!43!tablemid}{28.61} & \metriccell{tablepoor!71!tablemid}{30.79}\\
\mbox{CogVideoX1.5-5B~\cite{yang2025cogvideox}} & \metriccell{tablepoor!100!tablemid}{15.97} & \metriccell{tablepoor!100!tablemid}{28.74} & \metriccell{tablepoor!100!tablemid}{14.84} & \metriccell{tablepoor!100!tablemid}{12.35} & \metriccell{tablepoor!100!tablemid}{17.97}\\
\mbox{Cosmos3-Nano~\cite{agarwal2026cosmos}} & \metriccell{tablepoor!43!tablemid}{26.34} & \metriccell{tablepoor!43!tablemid}{41.74} & \metriccell{tablepoor!43!tablemid}{29.03} & \metriccell{tablepoor!14!tablemid}{35.18} & \metriccell{tablepoor!43!tablemid}{33.07}\\
\mbox{HunyuanVideo-1.5~\cite{wu2025hunyuanvideo}} & \metriccell{tablegood!71!tablemid}{30.45} & \metriccell{tablepoor!14!tablemid}{51.16} & \metriccell{tablepoor!14!tablemid}{35.42} & \metriccell{tablepoor!71!tablemid}{22.30} & \metriccell{tablepoor!14!tablemid}{34.83}\\
\mbox{MiniMax-H3~\cite{minimaxh3}} & \metriccell{tablepoor!71!tablemid}{26.09} & \metriccell{tablegood!14!tablemid}{55.53} & \metriccell{tablegood!14!tablemid}{41.83} & \metriccell{tablegood!14!tablemid}{38.90} & \metriccell{tablegood!14!tablemid}{40.59}\\
\mbox{GeoPhys~\cite{interno2026geophys}} & \metriccell{tablegood!14!tablemid}{29.49} & \metriccell{tablegood!71!tablemid}{57.04} & \metriccell{tablegood!71!tablemid}{43.88} & \metriccell{tablegood!43!tablemid}{40.06} & \metriccell{tablegood!43!tablemid}{42.62}\\
\mbox{WMReward~\cite{yuan2026inference}} & \metriccell{tablegood!43!tablemid}{30.13} & \metriccell{tablegood!43!tablemid}{56.72} & \metriccell{tablegood!43!tablemid}{43.69} & \metriccell{tablegood!71!tablemid}{40.99} & \metriccell{tablegood!71!tablemid}{\underline{42.88}}\\
\mbox{SPAR (ours)} & \metriccell{tablegood!100!tablemid}{30.57} & \metriccell{tablegood!100!tablemid}{59.89} & \metriccell{tablegood!100!tablemid}{46.05} & \metriccell{tablegood!100!tablemid}{41.39} & \metriccell{tablegood!100!tablemid}{\textbf{44.47}}\\
\bottomrule
\end{tabular*}
\end{minipage}
\end{table*}

\subsection{Experimental Setup}

\paragraph{Benchmarks.}
We evaluate image-to-video (I2V) generation on Physics-IQ Verified~\cite{radsch2026physics} and text-to-video (T2V) generation on PhyGenBench~\cite{meng2024towards} and VideoPhy-2 Hard~\cite{bansal2026videophy}.
Physics-IQ provides a fixed first frame for each of 198 inputs spanning 66 scenes and three views per scene.
PhyGenBench contains 160 prompts covering five physical categories, and VideoPhy-2 Hard contains 180 prompts from the hard subset.
For Physics-IQ, we report the official Verified score on a 0--100 scale, measuring continuation similarity to the reference video.
For PhyGenBench, we report the mean score from the open-source branch of PhyGenEval, multiplied by 100 and reported on a 0--100 scale throughout the paper.
For VideoPhy-2 Hard, we use GPT-6 Sol with the official prompts to assess semantic adherence (SA) and physical commonsense (PC) on five-point scales.
The primary metric is the percentage of videos scoring at least four on both criteria.
Evaluator settings and alternative-evaluator results are provided in Appendix~\ref{sec:additional-evaluators}.
\vspace{-1em}
\paragraph{Baselines and best-of-$N$ selection.}
We compare SPAR with WMReward~\cite{yuan2026inference} and GeoPhys~\cite{interno2026geophys}, using the same three MiniMax-H3~\cite{minimaxh3} candidates per input for best-of-$N$ selection ($N=3$).
The candidate pool contains 1,614 videos across 538 inputs.
The MiniMax-H3 baseline reports the exact expected score of uniformly selecting one of the three candidates per input, without reward-based selection.
We also evaluate video generation models including Wan2.2-TI2V-5B~\cite{wan2025wan}, CogVideoX1.5-5B~\cite{yang2025cogvideox}, Cosmos3-Nano~\cite{agarwal2026cosmos}, MinimaX-H3~\cite{minimaxh3} and HunyuanVideo-1.5~\cite{wu2025hunyuanvideo}.

\vspace{-1em}
\paragraph{Implementation details.}
\label{sec:implementation}
We use the frozen ViT-g V-JEPA~2~\cite{assran2025v} encoder and predictor.
Each video is sampled to 49 frames at $256\times256$ resolution and divided into five overlapping 16-frame windows with stride eight.
Each window uses eight history frames to predict the next eight frames, with the short-history prediction retaining the last four observed frames.
Static references repeat the first and middle sampled frames.
We use one configuration across all three benchmarks.
Spatial focusing selects the 64 highest-error positions, and temporal compensation uses the two highest-error windows.
The fusion coefficients are $\alpha=0.675$, $\lambda=0.25$, $\beta=0.60$, and $\gamma=-0.06875$.
Further implementation details appear in Appendix~\ref{sec:supp-implementation}, and coefficient sensitivity are analyzed in Appendix~\ref{sec:fusion-search}.

\subsection{Text-to-Video Physical Alignment}
\label{sec:t2v-results}

\begin{figure*}[t]
\centering
\includegraphics[width=\linewidth]{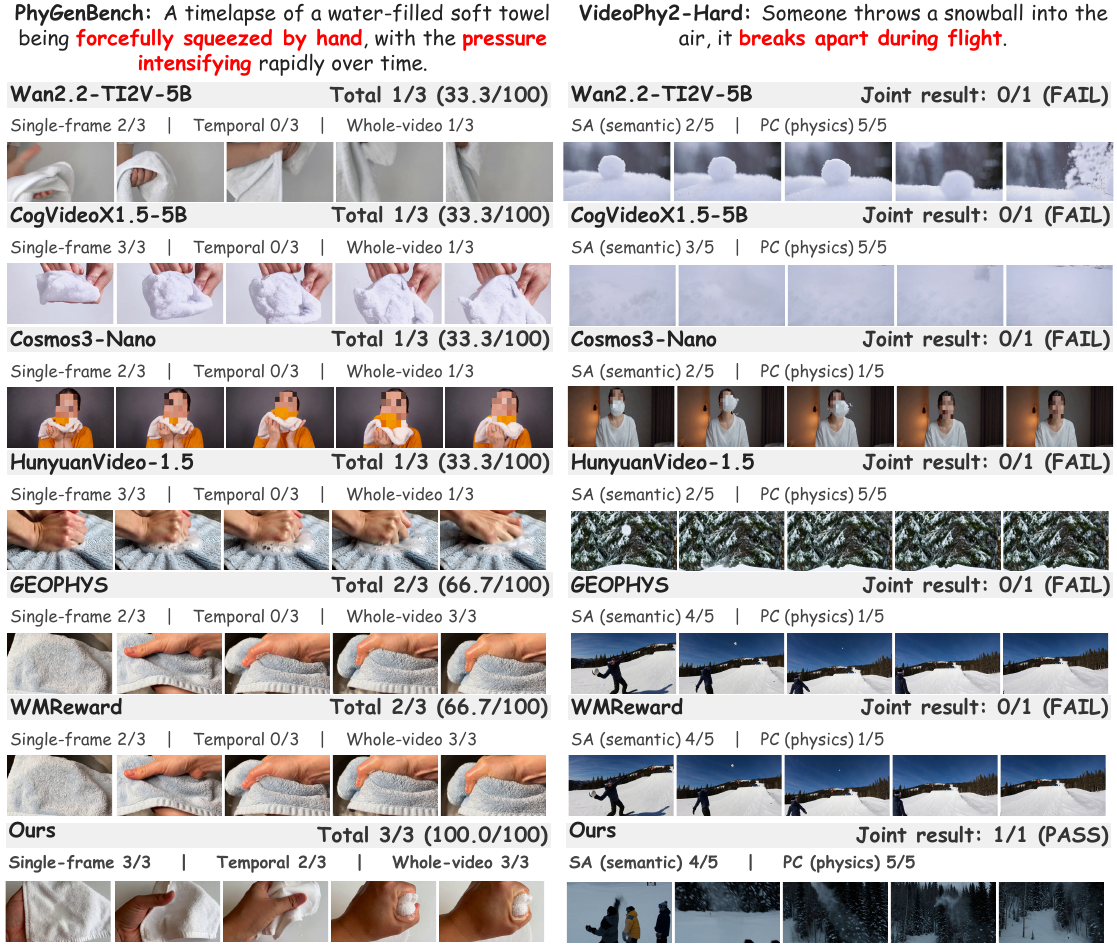}
\caption{\textbf{Text-to-video comparisons.} Results on PhyGenBench (left) and VideoPhy-2 Hard (right). Prompts are shown at the top, with key event requirements highlighted in red. Scores report each benchmark's overall evaluation and component scores.}
\label{fig:qualitative-t2v}
\end{figure*}

Text-conditioned generation must realize both the requested event and its physical progression.
Table~\ref{tab:main} shows that SPAR achieves competitive performance across all benchmarks.
Each entry also reports SPAR's pairwise win rate against the row method excluding ties.
Tables~\ref{tab:pgdetail} and~\ref{tab:vpdetail} provide detailed results on PhyGenBench and VideoPhy-2 Hard, respectively, with higher values indicating better performance.
Colors indicate within-column rank, and bold and underlined values mark the best and second-best overall scores, respectively.
Table~\ref{tab:pgdetail} shows gains over WMReward in four physical categories and a matching light score.
Table~\ref{tab:vpdetail} further shows that SPAR achieves higher joint semantic and physical success than both selection baselines.

The towel example in Figure~\ref{fig:qualitative-t2v} shows why the presence of the requested objects alone is insufficient.
WMReward and GeoPhys select a clip dominated by a hand pressing a towel, whereas the SPAR selection progresses from holding the towel to visibly compressing it in a closed grip, with water appearing below the fabric.
The snowball example tests whether the generated sequence captures the requested breakup during flight.
WMReward and GeoPhys select the same throwing sequence, whose displayed frames do not clearly show the breakup.
The SPAR selection shows a throw followed by dispersed snow against the trees, making the requested transition visible.
Appendix~\ref{sec:additional-evaluators} provides detailed evaluation protocols and scoring rules, along with additional results using GPT-6 Sol for PhyGenBench and AutoEval for VideoPhy-2 Hard.

\subsection{Image-to-Video Physical Alignment}
\label{sec:i2v-results}

\begin{figure}[t]
\centering
\includegraphics[width=\linewidth]{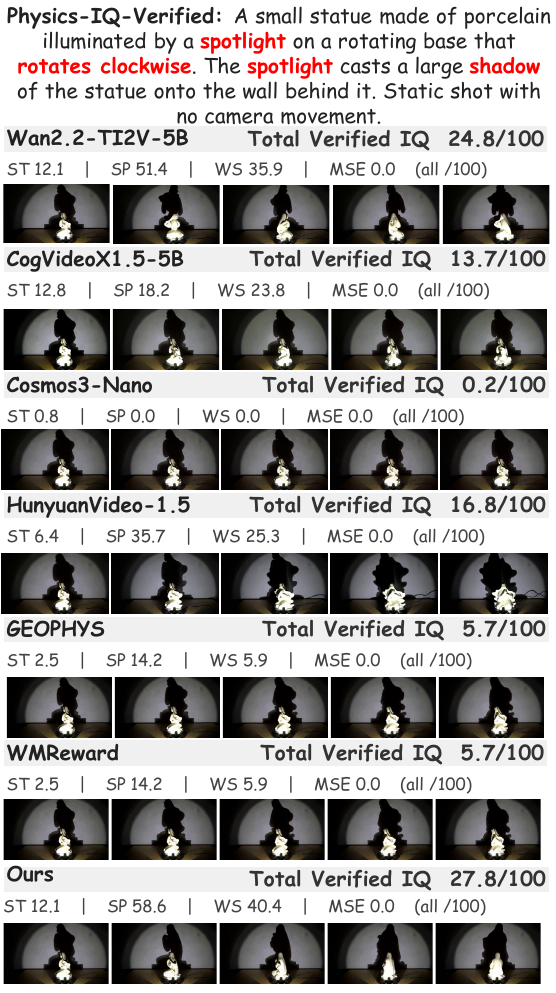}
\caption{\textbf{Image-to-video comparison on Physics-IQ.} SPAR selects smooth clockwise statue rotation with a changing shadow, while WMReward and GeoPhys select a largely static continuation. Scores are benchmark evaluations.}
\label{fig:qualitative-i2v}
\end{figure}

With a fixed initial frame, Physics-IQ tests how well generation continues the observed scene.
Table~\ref{tab:iqdetail} shows that SPAR improves all four continuation-similarity components over both selectors, with the largest gain over WMReward in spatial similarity.
Figure~\ref{fig:qualitative-i2v} illustrates the importance of evaluating evolution beyond the shared first frame.
The scene requires clockwise statue rotation under a fixed spotlight, with a corresponding change in the projected shadow.
The SPAR selection achieves smooth clockwise rotation accompanied by a changing shadow.
Neither WMReward nor GeoPhys selects a continuation that realizes this rotation, instead favoring a largely static statue.
This example illustrates how candidate selection can favor the requested evolution over a visually similar but largely static continuation.

\subsection{Ablation Studies}
\label{sec:ablation}

We evaluate SPAR's key designs through two complementary studies.
The combination study tests whether spatial and temporal refinement improve the basic self-referenced reward, whereas component ablations test whether each design remains useful when the other components are present.
Table~\ref{tab:ablation} reports module combinations and component ablations on Physics-IQ Verified (IQ), PhyGenBench (PG), and VideoPhy-2 Hard (VP).
Random selection reports the exact expected score of uniformly selecting one of the three MiniMax-H3 candidates per input.
SPAR-base selects candidates using only the difference between static-reference and real-video mean prediction errors.
Full adds both spatial and temporal refinement.
All configurations use the main comparison's candidate pool and evaluation protocols, retaining Full's coefficients for the active components.

\begin{table}[t]
\centering
\caption{\textbf{Module combinations and component ablations.} The first group compares random selection with the basic self-referenced reward (SPAR-base) and its spatial or temporal refinements. The second group removes or replaces individual components of the complete method (Full). Higher scores indicate better performance on all three benchmarks.}
\label{tab:ablation}
\small
\renewcommand{\arraystretch}{1.12}
\setlength{\tabcolsep}{3pt}
\begin{tabular*}{\linewidth}{@{}>{\raggedright\arraybackslash}p{\dimexpr.61\linewidth-\tabcolsep\relax}*{3}{>{\raggedleft\arraybackslash}p{\dimexpr.13\linewidth-2\tabcolsep\relax}}@{\hspace{\tabcolsep}}}
\toprule
\textbf{Configuration} & \textbf{IQ} & \textbf{PG} & \shortstack[r]{\textbf{VP}}\\
\midrule
\multicolumn{4}{@{}l}{\emph{Module combinations}}\\
Random selection & 40.59 & 62.08 & 59.81\\
SPAR-base & 43.49 & 64.17 & 60.56\\
SPAR-base + Spatial & 43.81 & 62.29 & 61.11\\
SPAR-base + Temporal & 43.71 & 64.58 & 60.56\\
\midrule
\multicolumn{4}{@{}l}{\emph{Component ablations of Full}}\\
W/o static references & 43.85 & 62.08 & 61.11\\
First-frame reference only & 43.24 & 64.58 & 59.44\\
Middle-frame reference only & 43.18 & 64.17 & 61.67\\
Hard-only weighting ($\lambda=0$) & 44.22 & 63.96 & 62.78\\
Continuous-only weighting ($\lambda=1$) & 43.88 & 64.58 & 62.22\\
W/o peak compensation ($\beta=0$) & 43.78 & 63.33 & 61.11\\
W/o history contrast ($\gamma=0$) & 44.29 & 62.71 & 62.22\\
\midrule
\rowcolor{tablegood!35}
\textbf{Full} & \textbf{44.47} & \textbf{65.00} & \textbf{62.78}\\
\bottomrule
\end{tabular*}
\end{table}
\vspace{-1em}
\paragraph{Module combinations.}
We evaluate spatial and temporal refinement separately and together to assess the benefit of their joint use.
Table~\ref{tab:ablation} shows that SPAR-base improves over random selection on all three benchmarks, while Full achieves the strongest results among the module combinations.
The advantage of Full over either separate refinement supports combining spatial and temporal evidence within the self-referenced reward.

\vspace{-1em}
\paragraph{Component ablations.}
To assess the specific choices within Full, we remove or replace its reference construction, spatial weighting, and temporal operations one at a time.
Table~\ref{tab:ablation} shows that removing static references or retaining only one anchor lowers performance on all three benchmarks, supporting the dual-anchor reference.
Hybrid spatial weighting outperforms either weighting scheme alone on IQ and PG and matches hard-only weighting on VP.
Removing either peak compensation or history contrast also lowers performance across all three benchmarks, supporting the contribution of each temporal operation.
Further ablations on parameter sensitivity appear in Appendix~\ref{sec:full-ablation}.

\section{Conclusion}
\label{sec:conclusion}

We presented SPAR, a framework for inference-time physical alignment using self-referenced world-model rewards.
Static references constructed from each video's own frames provide a content-matched baseline for evaluating its dynamics, while spatiotemporal reward refinement incorporates localized discrepancies and temporal changes into the final reward.
Evaluations across multiple benchmarks demonstrate improved physical alignment in both text-to-video generation and image-to-video continuation.
These results highlight the value of interpreting world-model predictions relative to each video's own content when evaluating generated dynamics.

\paragraph{Limitations.}
SPAR currently improves physical alignment through candidate selection, so its gains depend on the quality and diversity of the generated candidates.
It cannot correct physical errors shared by all candidates.
Extending self-referenced rewards to guide generation directly is a promising direction for future work.

\clearpage
{
\small
\bibliographystyle{ieeenat_fullname}
\bibliography{references}
}
\clearpage
\appendix
\onecolumn
\setcounter{table}{0}
\providecommand{\mainpaperref}[1]{\ref{#1}}
\setcounter{figure}{0}
\renewcommand{\thetable}{S\arabic{table}}
\renewcommand{\thefigure}{S\arabic{figure}}
\renewcommand{\theHtable}{supp.\arabic{table}}
\renewcommand{\theHfigure}{supp.\arabic{figure}}
\tableformat
\suppressfloats[t]
\begin{center}
{\LARGE\bfseries SPAR: Self-Referenced World-Model Rewards for Physical Alignment in Video Generation\par}
\vspace{0.3cm}
{\LARGE\bfseries Supplementary Material\par}
\end{center}
\vspace{0.5cm}

This supplementary material presents background and theoretical analysis (Appendices~\ref{sec:background} and~\ref{sec:supp-method}), followed by evaluation protocols and additional results (Appendix~\ref{sec:additional-evaluators}). These include \textbf{human evaluation of physical plausibility and event completion}, \textbf{reward transfer to Wan with the same configuration used for MiniMax-H3}, and \textbf{best-of-$N$ scaling with candidate budgets from 1 to 7}. Analyses of reward components and robustness are presented in Appendix~\ref{sec:full-ablation}.

\section{Background}
\label{sec:background}

This section reviews physical knowledge in video generation, inference-time alignment, and physical evaluation.

\subsection{Training-Based Physical Video Generation}

Physical supervision can be introduced through specialized examples and annotations.
PISA~\cite{li2025pisa} studies post-training on simulated freefall videos, while PhyCo~\cite{narayanan2026phyco} combines simulation videos, physical-property-map conditioning, and VLM-guided reward optimization.
Physis-Lang~\cite{lu2026physis} uses physical descriptions to connect targeted data retrieval, generator training, and inference-time prompting.
It therefore incorporates physical knowledge during both training and inference.

Other methods learn representations or objectives that make physical information available to the generator.
MMPhysVideo~\cite{lin2026mmphysvideo} jointly models RGB and perceptual cues before distilling their representation into a generator.
NewtonGen~\cite{yuan2026newtongen} introduces learned Newtonian dynamics, and MotionRAG~\cite{zhu2026motionrag} trains modules that adapt motion from retrieved reference videos and inject it into generation.
PhysCorr~\cite{wang2025physcorr} uses physical reward and preference optimization, while Diffusion-DRF~\cite{wang2026diffusion} fine-tunes generators with feedback from a frozen VLM.
Across these training routes, the additional requirements may involve simulation, caption generation, perceptual annotations, reference-video organization, or optimization of auxiliary modules.
SPAR uses the video's own static content to calibrate a frozen world model's reward for inference-time physical alignment, without constructing a new training corpus or updating network weights.

\subsection{Inference-Time Physical Alignment}

Inference-time alignment keeps generator weights fixed while modifying input conditions, sampling, or output selection.
At the input level, PhyT2V~\cite{xue2025phyt2v} and VLM-guided self-refinement~\cite{liu2025bootstrapping} iteratively revise prompts using reasoning or feedback about generated videos.
PhyPrompt~\cite{wu2026phyprompt} applies a physics-aware prompt refiner that is trained separately from the video generator.
At the sampling level, implausibility-aware guidance~\cite{hao2025enhancing} uses negative descriptions to steer denoising, and CausalMotion~\cite{zhuang2026causalmotion} supplies planned keyframes and object trajectories.
The physical consistency achieved by these approaches depends on the generator's fidelity to the supplied conditions, and structured guidance requires compatible generation interfaces.

Reward-based methods use a score to guide sampling or select outputs.
Feynman--Kac steering~\cite{singhal2025general} provides a general framework for reward-guided diffusion sampling.
Seeking Physics in Diffusion Noise~\cite{tang2026seeking} trains a lightweight verifier on frozen denoising features and uses it for trajectory selection or reward-gradient guidance.
Although the generator remains fixed, this route requires backbone-specific verifier training and access to intermediate generator states.
Best-of-$N$ search instead ranks independent samples from the generator using an external physical reward, aligning generation toward higher-reward outcomes without accessing intermediate denoising states.

GeoPhys~\cite{interno2026geophys} and WMReward~\cite{yuan2026inference} illustrate two sources of external physical evaluation signals.
GeoPhys aggregates geometric statistics of feature trajectories from frozen image encoders, including speed variation, curvature, acceleration, and a linear prediction residual.
WMReward uses the encoder and predictor of V-JEPA~2~\cite{assran2025v} to measure disagreement between predicted and observed future representations.
It uses this predictive signal for both selection and generation guidance.
SPAR implements self-referenced physical alignment through reward-driven best-of-$N$ search, with both the generator and world model fixed.
Its static references establish a baseline matched to each video's content before the prediction statistics are combined into a reward.

\subsection{Physical Understanding and Video Evaluation}

Violation-of-expectation evaluation tests whether models respond differently to physically possible and impossible events.
IntPhys~2~\cite{bordes2025intphys} provides controlled tests of object permanence, immutability, spatiotemporal continuity, and solidity.
Garrido \etal~\cite{garrido2025intuitive} show that self-supervised prediction in a learned representation space can yield sensitivity to violations of object permanence and shape consistency.
These studies provide evidence for using predictive representations in physical evaluation, while generated-video benchmarks test whether the resulting reward improves actual outputs.

Physics-IQ~\cite{motamed2026generative} evaluates continuations of real physical events, and Physics-IQ Verified~\cite{radsch2026physics} revises its reference data and evaluation workflow.
PhyGenBench~\cite{meng2024towards} covers diverse physical phenomena with a hierarchical evaluation pipeline.
VideoPhy-2~\cite{bansal2026videophy} focuses on action-centric events and separates semantic adherence from physical commonsense.
These protocols emphasize complementary properties: fidelity to observed evolution, consistency with physical principles, and fulfillment of the requested event.
Our experiments use their evaluation scores to assess selected videos independently of the world-model reward.
The main paper specifies the primary evaluation protocols, and Appendix~\ref{sec:additional-evaluators} reports the alternative evaluators and their aggregation rules.

\section{Method Analysis and Configuration}
\label{sec:supp-method}

This section analyzes the self-referenced reward decomposition (Section~\ref{sec:supp-reward-decomposition}), derives the spatial weight allocation (Section~\ref{sec:supp-spatial-allocation}), and explains the temporal refinement terms (Section~\ref{sec:supp-temporal-refinement}).
It also specifies feature extraction and the shared reward configuration (Section~\ref{sec:supp-implementation}).

\subsection{Self-Referenced Reward Decomposition}
\label{sec:supp-reward-decomposition}
For candidate $v_i$, write $B_i=B(v_i)$, $E_{\mathrm r,i}=\langle E\rangle_{\mathrm r,i}$, and $E_{\mathrm f,i}=\langle E\rangle_{\mathrm f,i}$, where the additional subscript $i$ identifies the candidate whose windows and references are averaged.
Then
\begin{equation}
B_i-B_j=-(E_{\mathrm r,i}-E_{\mathrm r,j})+(E_{\mathrm f,i}-E_{\mathrm f,j}).
\label{eq:cancel}
\end{equation}
A shared reference cancels from this difference.
If scene content contributes the same additive term $\kappa_i$ to both errors, we can write $E_{\mathrm r,i}=\kappa_i+\delta_i$ and $E_{\mathrm f,i}=\kappa_i+\epsilon_i$, where $\delta_i$ and $\epsilon_i$ are the real-video and static-reference residuals.
Under this assumption, the shared term cancels, giving $B_i=\epsilon_i-\delta_i$.
This identity explains how a matched reference can remove a shared additive contribution.
For a fixed candidate, define $H=\langle h\rangle_{\mathrm f}-\langle h\rangle_{\mathrm r}$, $S=\langle s\rangle_{\mathrm f}-\langle s\rangle_{\mathrm r}$, and $T=(1-\lambda)H+\lambda S$, with the argument $v$ omitted from video-level statistics for brevity.
Substituting $q_w$ into Eq.~\mainpaperref{eq:reward} yields $R=B+\alpha(T-B)+\beta P+\gamma D$.
Equivalently, defining $\phi_w=q_w-\gamma d_w$ yields $R=\langle\phi\rangle_{\mathrm f}-\langle\phi\rangle_{\mathrm r}+\beta P$.

\subsection{Spatial Weight Allocation}
\label{sec:supp-spatial-allocation}
For $c\geq1$, the continuous weights solve the constrained optimization problem
\begin{equation}
\max_{0\leq\omega_w(p)\leq c,\,\operatorname{mean}_p\omega_w(p)=1}
\frac1{P_s}\sum_p\left[\omega_w(p)u_w(p)-\frac12\omega_w(p)^2\right].
\label{eq:softobjective}
\end{equation}
The stationarity conditions with a multiplier for the mean constraint yield $\omega_w(p)=\min\{c,\max\{0,u_w(p)-\tau_w\}\}$.
The quadratic term yields a graded allocation of the fixed total weight.
The strictly concave quadratic objective has a unique optimum, subject to the stated constraints.
Without the quadratic term, setting $c=P_s/K$ yields an optimum that assigns weight $c$ to the $K$ highest-error positions and zero elsewhere.
Equal-valued positions at the cutoff can share weight without changing the optimal statistic.
The two branches differ in how sharply they allocate a common spatial weight budget.

For each standardized error map, the mean clipped weight is a continuous, nonincreasing function of $\tau_w$.
The interval $[\min_p u_w(p)-c,\max_p u_w(p)]$ brackets a solution to the unit-mean constraint, since its endpoints give mean weights $c$ and zero, respectively.
For a constant error map, the standard deviation floor makes every standardized value zero; the normalized weights become one and the weighted mean equals the ordinary mean.
Thus, continuous focusing reduces to uniform averaging when the map contains no spatial contrast.

\subsection{Temporal Reward Refinement}
\label{sec:supp-temporal-refinement}

\paragraph{Peak compensation as temporal reweighting.}
Separating the mean error $E_w$ from the additional spatial contribution $q_w-E_w$, the full reward can be rewritten as
\begin{align}
R={}&\langle q\rangle_{\mathrm{f}}
-\big[(1+\beta)\langle E\rangle_{\mathrm{r}}
-\beta\operatorname{TopMean}_{m}(E)\big]\nonumber\\
&-\langle q-E\rangle_{\mathrm{r}}+\gamma D.
\label{eq:temporal-reweighting}
\end{align}
For $W$ real windows and $1\leq m<W$ peaks, the bracketed mean-error penalty assigns weight $(1+\beta)/W-\beta/m$ to each peak and $(1+\beta)/W$ to each remaining window.
These weights sum to one and are nonnegative for $0\leq\beta\leq m/(W-m)$.
The default $W=5$, $m=2$, and $\beta=.60$ satisfy this condition and yield weights $.02$ and $.32$, respectively.

\paragraph{History-conditioned directional agreement.}
For history $h\in\{\ell,s\}$, define $\Delta z_{w,t,p}=z_{w,t+1,p}-z_{w,t,p}$ and $\Delta\hat z^h_{w,t,p}=\hat z^h_{w,t+1,p}-\hat z^h_{w,t,p}$ for $t=1,\ldots,T_f-1$.
Using the normalized features from the main paper, let
\begin{align}
    U_{w,p}&=\operatorname{mean}_{t}\|\Delta z_{w,t,p}\|^2,\\
    V^{h}_{w,p}&=\operatorname{mean}_{t}\|\Delta\hat{z}^{h}_{w,t,p}\|^2,\\
    J^{h}_{w,p}&=\operatorname{mean}_{t}\|\Delta z_{w,t,p}-\Delta\hat{z}^{h}_{w,t,p}\|^2.
\end{align}
Here $\operatorname{mean}_t$ averages the $T_f-1$ consecutive differences; $T_f=4$ gives three differences per spatial position.
The normalized directional agreement is
\begin{equation}
    a^{h}_{w,p}=\frac{U_{w,p}+V^{h}_{w,p}-J^{h}_{w,p}}
    {2\sqrt{U_{w,p}V^{h}_{w,p}}+\epsilon_d}.
\end{equation}
The numerator is twice the mean inner product of the predicted and observed differences.
The denominator normalizes by their magnitudes, with $\epsilon_d=10^{-6}$.
Dividing numerator and denominator by two gives the main paper's normalized-inner-product expression with stabilizer $\epsilon_d/2$.
We average the difference in agreement between long and short histories over spatial positions and subtract the corresponding mean for the two static references, obtaining the history-conditioned directional contrast $D(v)$ defined in the main paper.
This measures feature-change alignment, not optical flow or physical velocity.

\subsection{Feature Extraction and Reward Configuration}
\label{sec:supp-implementation}

We use the pretrained ViT-g V-JEPA~2~\cite{assran2025v} encoder and predictor.
The predictor has 12 layers, 12 attention heads, embedding dimension 384, and rotary positional embeddings; the encoder uses spatial patch size 16 and temporal tubelet size two.
Both networks remain frozen during reward computation.
Full-window target features and context-masked encoder features use the same encoder weights.
The long- and short-history context masks retain four and two temporal tubelets, respectively, while both predict the four future tubelets at the same absolute positions.

For a source video containing $F_{\mathrm{src}}\geq F$ frames, we sample $F=49$ frames over its full duration.
With frame indices starting at zero, sampled index $j\in\{0,\ldots,F-1\}$ corresponds to original frame $\lfloor j(F_{\mathrm{src}}-1)/(F-1)\rfloor$.
RGB frames are bilinearly resized to $256\times256$, scaled to $[0,1]$, and normalized with channel means $(.485,.456,.406)$ and standard deviations $(.229,.224,.225)$.
Context masks are applied before encoder attention, and the short-history mask retains the absolute positions of the last two context tubelets.
The target features are shared between the long- and short-history predictions.

Feature computation uses FP32 and five windows of length $L=16$, starting at sampled indices $0,8,16,24,32$.
Sampled frame 48 is unused by these windows.
The long history and future each contain eight frames, and the short history retains the last four observed frames.
With tubelet size $b=2$ and spatial patch size 16, each window has $T_f=4$ future temporal positions and $P_s=256$ spatial positions.
The static anchors are sampled frames $a_1=0$ and $a_2=24$, using the anchor notation from the main paper.
Spatial focusing uses $K=64$, weight cap $c=4$, and 60 bisection iterations to determine the normalization threshold, with $\epsilon_s=10^{-12}$.
Temporal concentration uses $m=2$, and directional normalization uses $\epsilon_d=10^{-6}$.
The shared coefficients are $\alpha=.675$, $\lambda=.25$, $\beta=.60$, and $\gamma=-.06875$.
Exact reward ties are resolved by the original candidate order.

\section{Evaluation Protocols and Additional Results}
\label{sec:additional-evaluators}

This section describes the evaluation protocols (Sections~\ref{sec:benchmark-metrics}--\ref{sec:gpt-protocol}) and presents additional results on evaluator variation (Sections~\ref{sec:phygen-gpt}--\ref{sec:videophy-autoeval}), human assessment (Section~\ref{sec:supp-user-study}), reward transfer (Section~\ref{sec:supp-wan-transfer}), and candidate-budget scaling (Section~\ref{sec:supp-budget-scaling}).
Additional qualitative comparisons across the three benchmarks are presented in Figures~\ref{fig:supp-qualitative-1}--\ref{fig:supp-qualitative-7} at the end of the appendix.

\subsection{Benchmark Metrics and Reporting}
\label{sec:benchmark-metrics}

\paragraph{Physics-IQ Verified.}
The official evaluation measures continuation similarity using four components, each clipped to $[0,1]$ for each scene and view before aggregation.
We report scores on a 0--100 scale; the overall score equals the mean of the four unrounded component scores.
The MSE component converts mean squared error into a normalized, inverted, and clipped similarity score, so higher values indicate better performance.
\paragraph{PhyGenBench.}
The main comparison uses the open-source branch of PhyGenEval.
Its overall and category scores are multiplied by 100 and reported on a 0--100 scale.
The benchmark contains 40 Force, 50 Light, 30 Heat, 25 Physical Properties, and 15 Chemical Properties inputs.
The overall score is the mean over all 160 inputs, weighting each category in proportion to its number of inputs.
\paragraph{VideoPhy-2 Hard.}
Semantic adherence (SA) and physical commonsense (PC) are assessed on five-point scales.
We report their mean scores, the percentage of videos scoring at least four on each criterion, and the percentage satisfying both thresholds (Overall).
Joint success therefore requires both criteria to be satisfied by the same video.

\subsection{GPT-6 Sol Evaluation Protocol}
\label{sec:gpt-protocol}

GPT-6 Sol evaluation uses low reasoning effort, low image detail, and an output limit of 2,048 tokens. The model is accessed through a third-party service under the GPT-6 Sol identifier.
This evaluator provides the main VideoPhy-2 Hard results and the additional PhyGenBench results in Section~\ref{sec:phygen-gpt}.
When comparing evaluators, we hold candidate selections and reward parameters fixed.
For the automatic evaluations, the MiniMax-H3 baseline is the mean score over all three candidates per input, equivalent to the expected score under uniform random selection.

For VideoPhy-2 Hard, we sample 32 frames at segment centers, resize them to $224\times224$, and assess semantic adherence (SA) and physical commonsense (PC) separately using the official prompts.
The 180 inputs are scored using the criteria and joint-success threshold defined in Section~\ref{sec:benchmark-metrics}.

\subsection{PhyGenBench with GPT-6 Sol}
\label{sec:phygen-gpt}

To examine sensitivity to the evaluator, we additionally evaluate PhyGenBench using GPT-6 Sol within the PhyGenEval three-stage framework.
We retain the VQAScore key-phenomenon scores and retrieval results, and replace the evaluators for the temporal and whole-video stages with GPT-6 Sol.
Let $y_n^{\mathrm{single}}$, $y_n^{\mathrm{multi}}$, and $y_n^{\mathrm{video}}$ denote the stage scores on a 0--3 scale for the selected video of input $n$, assessing the key phenomenon, temporal progression, and whole-video naturalness, respectively.
Following the official closed-branch aggregation, the per-video score is
\begin{equation}
y_n^{\mathrm{PG}}=\frac{100}{3}\operatorname{round}
\left(\frac{y_n^{\mathrm{single}}+y_n^{\mathrm{multi}}+y_n^{\mathrm{video}}}{3}\right).
\end{equation}
The overall score averages $y_n^{\mathrm{PG}}$ over all 160 inputs, with rounding applied to each video's mean stage score before averaging across inputs.
For category reporting, Material combines the 25 Physical Properties and 15 Chemical Properties inputs, weighting their category means by $25/40$ and $15/40$, respectively.
The overall score remains the mean over inputs.

\begin{table*}[t]
\centering\footnotesize\tableformat\setlength{\tabcolsep}{3pt}
\caption{\textbf{PhyGenBench with GPT-6 Sol.} GPT-6 Sol replaces the temporal and whole-video evaluators within the three-stage framework; scores follow the official closed-branch aggregation on a 0--100 scale. Colors rank methods within each column. Bold and underline mark the best and second-best overall scores.}
\label{tab:phygen-gpt}
\begin{tabular*}{\textwidth}{@{}>{\raggedright\arraybackslash}p{\dimexpr.30\textwidth-\tabcolsep\relax}*{5}{>{\raggedleft\arraybackslash}p{\dimexpr.14\textwidth-2\tabcolsep\relax}}@{\hspace{\tabcolsep}}}
\toprule
{\bfseries\boldmath Method} & {\boldmath Force} & {\boldmath Light} & {\boldmath Heat} & {\boldmath Material} & {\bfseries\boldmath Overall}\\
\midrule
Wan2.2-TI2V-5B~\cite{wan2025wan} & \metriccell{tablepoor!20!tablemid}{45.83} & \metriccell{tablepoor!33!tablemid}{62.00} & \metriccell{tablepoor!60!tablemid}{41.11} & \metriccell{tablepoor!67!tablemid}{38.33} & \metriccell{tablepoor!43!tablemid}{48.13}\\
CogVideoX1.5-5B~\cite{yang2025cogvideox} & \metriccell{tablepoor!60!tablemid}{43.33} & \metriccell{tablepoor!67!tablemid}{55.33} & \metriccell{tablegood!100!tablemid}{48.89} & \metriccell{tablepoor!33!tablemid}{42.50} & \metriccell{tablepoor!71!tablemid}{47.92}\\
Cosmos3-Nano~\cite{agarwal2026cosmos} & \metriccell{tablepoor!100!tablemid}{40.00} & \metriccell{tablepoor!100!tablemid}{51.33} & \metriccell{tablepoor!100!tablemid}{33.33} & \metriccell{tablepoor!100!tablemid}{36.67} & \metriccell{tablepoor!100!tablemid}{41.46}\\
HunyuanVideo-1.5~\cite{wu2025hunyuanvideo} & \metriccell{tablegood!100!tablemid}{53.33} & \metriccell{tablegood!33!tablemid}{66.00} & \metriccell{tablegood!100!tablemid}{48.89} & \metriccell{tablepoor!33!tablemid}{42.50} & \metriccell{tablepoor!14!tablemid}{53.75}\\
MiniMax-H3~\cite{minimaxh3} & \metriccell{tablegood!20!tablemid}{51.67} & \metriccell{tablegood!67!tablemid}{66.22} & \metriccell{tablegood!60!tablemid}{44.44} & \metriccell{tablegood!0!tablemid}{49.44} & \metriccell{tablegood!43!tablemid}{54.31}\\
GeoPhys~\cite{interno2026geophys} & \metriccell{tablegood!20!tablemid}{51.67} & \metriccell{tablegood!100!tablemid}{69.33} & \metriccell{tablepoor!20!tablemid}{42.22} & \metriccell{tablegood!67!tablemid}{52.50} & \metriccell{tablegood!100!tablemid}{\textbf{55.63}}\\
WMReward~\cite{yuan2026inference} & \metriccell{tablegood!100!tablemid}{53.33} & \metriccell{tablegood!0!tablemid}{65.33} & \metriccell{tablepoor!20!tablemid}{42.22} & \metriccell{tablegood!33!tablemid}{50.00} & \metriccell{tablegood!14!tablemid}{54.17}\\
SPAR (ours) & \metriccell{tablegood!60!tablemid}{52.50} & \metriccell{tablegood!33!tablemid}{66.00} & \metriccell{tablegood!20!tablemid}{43.33} & \metriccell{tablegood!100!tablemid}{53.33} & \metriccell{tablegood!71!tablemid}{\underline{55.21}}\\
\bottomrule
\end{tabular*}
\end{table*}

As shown in Table~\ref{tab:phygen-gpt}, SPAR achieves an overall score of 55.21, compared with 54.17 for WMReward and 55.63 for GeoPhys.
SPAR obtains the highest Material score, 53.33, and improves over WMReward under both this protocol and the open-source evaluation in the main paper.

\subsection{VideoPhy-2 Hard with AutoEval}
\label{sec:videophy-autoeval}

We evaluate the same selected videos on the 180 VideoPhy-2 Hard inputs using AutoEval, retaining the scoring scales and success thresholds in Section~\ref{sec:benchmark-metrics}.
Table~\ref{tab:vp-autoeval} provides a comparison with the GPT-6 Sol evaluation used in the main paper.

\begin{table*}[t]
\centering\footnotesize\tableformat\setlength{\tabcolsep}{3pt}
\caption{\textbf{VideoPhy-2 Hard with AutoEval.} SA/PC are mean scores. Pass rates and overall joint success are percentages.}
\label{tab:vp-autoeval}
\begin{tabular*}{\textwidth}{@{}>{\raggedright\arraybackslash}p{\dimexpr.30\textwidth-\tabcolsep\relax}*{5}{>{\raggedleft\arraybackslash}p{\dimexpr.14\textwidth-2\tabcolsep\relax}}@{\hspace{\tabcolsep}}}
\toprule
{\bfseries\boldmath Method} & {\boldmath SA} & {\boldmath PC} & {\boldmath SA$\geq4$} & {\boldmath PC$\geq4$} & {\boldmath Overall}\\
\midrule
Wan2.2-TI2V-5B~\cite{wan2025wan} & \metriccell{tablegood!100!tablemid}{2.87} & \metriccell{tablepoor!33!tablemid}{3.47} & \metriccell{tablegood!20!tablemid}{12.22} & \metriccell{tablepoor!67!tablemid}{43.33} & \metriccell{tablegood!20!tablemid}{8.33}\\
CogVideoX1.5-5B~\cite{yang2025cogvideox} & \metriccell{tablegood!60!tablemid}{2.84} & \metriccell{tablepoor!67!tablemid}{3.46} & \metriccell{tablegood!100!tablemid}{13.33} & \metriccell{tablepoor!33!tablemid}{43.89} & \metriccell{tablegood!100!tablemid}{\textbf{10.56}}\\
Cosmos3-Nano~\cite{agarwal2026cosmos} & \metriccell{tablepoor!100!tablemid}{2.59} & \metriccell{tablegood!100!tablemid}{3.72} & \metriccell{tablepoor!100!tablemid}{7.78} & \metriccell{tablegood!100!tablemid}{60.00} & \metriccell{tablepoor!100!tablemid}{5.56}\\
HunyuanVideo-1.5~\cite{wu2025hunyuanvideo} & \metriccell{tablegood!60!tablemid}{2.84} & \metriccell{tablepoor!100!tablemid}{3.38} & \metriccell{tablegood!60!tablemid}{12.78} & \metriccell{tablepoor!100!tablemid}{36.67} & \metriccell{tablegood!20!tablemid}{8.33}\\
MiniMax-H3~\cite{minimaxh3} & \metriccell{tablepoor!60!tablemid}{2.80} & \metriccell{tablegood!0!tablemid}{3.51} & \metriccell{tablepoor!60!tablemid}{10.93} & \metriccell{tablegood!33!tablemid}{46.11} & \metriccell{tablepoor!20!tablemid}{7.96}\\
GeoPhys~\cite{interno2026geophys} & \metriccell{tablegood!20!tablemid}{2.83} & \metriccell{tablegood!67!tablemid}{3.53} & \metriccell{tablegood!20!tablemid}{12.22} & \metriccell{tablegood!33!tablemid}{46.11} & \metriccell{tablegood!20!tablemid}{8.33}\\
WMReward~\cite{yuan2026inference} & \metriccell{tablepoor!20!tablemid}{2.81} & \metriccell{tablegood!67!tablemid}{3.53} & \metriccell{tablepoor!20!tablemid}{11.67} & \metriccell{tablegood!67!tablemid}{46.67} & \metriccell{tablepoor!60!tablemid}{7.22}\\
SPAR (ours) & \metriccell{tablepoor!20!tablemid}{2.81} & \metriccell{tablegood!33!tablemid}{3.51} & \metriccell{tablegood!20!tablemid}{12.22} & \metriccell{tablegood!0!tablemid}{45.00} & \metriccell{tablegood!60!tablemid}{\underline{9.44}}\\
\bottomrule
\end{tabular*}
\end{table*}

Under AutoEval, SPAR selects 17 jointly successful videos, compared with 13 for WMReward and 15 for GeoPhys.
GPT-6 Sol assigns joint success to 113, 110, and 105 videos for SPAR, WMReward, and GeoPhys, respectively.
Although the evaluators yield different absolute success rates, both rank SPAR above WMReward and GeoPhys on joint success.

\subsection{User Study}
\label{sec:supp-user-study}

\paragraph{Study design.}
Human evaluation assesses physical plausibility and event completion as separate properties of the generated videos.
The study covers 15 scenes, with five from each benchmark.
PhyGenBench contributes one randomly sampled input from each of its five physical categories; Physics-IQ contributes five distinct scenes with one randomly sampled view per scene; and VideoPhy-2 contributes five randomly sampled inputs from its Hard subset.
Scene sampling is independent of automatic scores and human ratings.
Each questionnaire contains six scenes, two per benchmark, with independently randomized scene and video orders.
We analyze the first 20 completed questionnaires, yielding eight evaluations per scene and 744 video evaluations after accounting for duplicate selections within each questionnaire.

The evaluated methods comprise Wan2.2-TI2V-5B, CogVideoX1.5-5B, Cosmos3-Nano, HunyuanVideo-1.5, MiniMax-H3, WMReward, GeoPhys, and SPAR.
The three reward methods select from the same three MiniMax-H3 candidates per input, while the MiniMax-H3 reference uses one uniformly sampled candidate fixed before rating.
Each of the four other generators contributes one video per input.
Method identities and automatic scores are hidden from raters.
Identical videos selected by multiple methods are presented once per scene, and their ratings are assigned to each corresponding method.
In this study, each scene contains six or seven distinct videos.
For each scene, raters are shown the event description and, for image-to-video generation, the initial conditioning frame. They may replay the videos as needed.

\paragraph{Rating criteria.}
Physical plausibility measures whether motion, contact, deformation, and material changes follow real-world physical behavior.
Raters are instructed to judge these dynamics independently of visual sharpness or aesthetic quality.
Event completion measures whether the main action and its outcome occur as described, distinguishing realized events from the mere presence of the relevant objects.
The two criteria are rated separately on five-point scales, from clearly implausible to highly plausible and from absent to fully realized, respectively.
An ``unable to judge'' response is treated as missing for the corresponding criterion.
Raters may assign equal scores to different videos.

\paragraph{Human ratings.}
Table~\ref{tab:supp-human-ratings} reports mean ratings for physical plausibility and event completion, with each criterion averaged over its valid responses.
SPAR receives the highest mean on both criteria, scoring 4.083 for physical plausibility and 4.317 for event completion.
WMReward obtains the second-highest means, 4.000 and 4.208, respectively.
These ratings support the physical plausibility and event completion of videos selected by SPAR.

\begin{table}[ht]
\centering
\caption{\textbf{Human evaluation across three benchmarks.} Both criteria use a 1--5 scale, with higher scores indicating better quality. Counts denote valid ratings per method; within each questionnaire, identical selected videos share the same rating. MiniMax-H3 uses one fixed uniformly sampled candidate per input. Bold and underline mark the best and second-best means.}
\label{tab:supp-human-ratings}
\small
\begin{tabular*}{\linewidth}{@{\extracolsep{\fill}}lrrrr@{}}
\toprule
& \multicolumn{2}{c}{Physical plausibility} & \multicolumn{2}{c}{Event completion}\\
\cmidrule(lr){2-3}\cmidrule(lr){4-5}
Method & Mean $\uparrow$ & Valid ratings & Mean $\uparrow$ & Valid ratings\\
\midrule
Wan2.2-TI2V-5B~\cite{wan2025wan} & 2.208 & 120 & 2.150 & 120\\
CogVideoX1.5-5B~\cite{yang2025cogvideox} & 1.664 & 119 & 2.083 & 120\\
Cosmos3-Nano~\cite{agarwal2026cosmos} & 2.416 & 113 & 2.289 & 114\\
HunyuanVideo-1.5~\cite{wu2025hunyuanvideo} & 2.538 & 119 & 2.658 & 120\\
\midrule
MiniMax-H3~\cite{minimaxh3} & 3.655 & 119 & 3.842 & 120\\
WMReward~\cite{yuan2026inference} & \underline{4.000} & 120 & \underline{4.208} & 120\\
GeoPhys~\cite{interno2026geophys} & 3.697 & 119 & 3.817 & 120\\
SPAR & \textbf{4.083} & 120 & \textbf{4.317} & 120\\
\bottomrule
\end{tabular*}
\end{table}

\subsection{Reward Transfer to Wan}
\label{sec:supp-wan-transfer}

We test whether SPAR transfers to a different generator by applying it to Wan2.2-TI2V-5B on PhyGenBench and VideoPhy-2 Hard.
All selectors rank the same three candidates per input ($N=3$), giving 480 candidates for the 160 PhyGenBench inputs and 540 for the 180 VideoPhy-2 inputs.
\textbf{SPAR uses exactly the same reward configuration as the MiniMax-H3 experiments in the main paper}, including $\alpha=.675$, $\lambda=.25$, $\beta=.60$, and $\gamma=-.06875$, with the feature extraction and reference construction specified in Appendix~\ref{sec:supp-implementation}.
Both benchmarks retain the main-paper evaluation protocols: PhyGenEval open for PhyGenBench and GPT-6 Sol for VideoPhy-2 Hard (Sections~\ref{sec:benchmark-metrics} and~\ref{sec:gpt-protocol}).

\begin{table}[ht]
\centering
\caption{\textbf{Reward transfer to Wan2.2-TI2V-5B~\cite{wan2025wan}.} All selectors use the same three candidates per input; random selection is the exact uniform expectation. PhyGenBench scores use a 0--100 scale; Chem. and Phys. denote chemical and physical properties. VideoPhy-2 reports mean semantic adherence (SA) and physical commonsense (PC) on a 1--5 scale, with marginal pass rates and joint success (Overall) in percent. Higher is better. Colors indicate within-column rank. Bold and underline mark the best and second-best overall scores.}
\label{tab:supp-wan-transfer}
\footnotesize
\tableformat
\setlength{\tabcolsep}{3pt}
\begin{tabular*}{\linewidth}{@{}>{\raggedright\arraybackslash}p{\dimexpr.18\linewidth-\tabcolsep\relax}*{11}{>{\raggedleft\arraybackslash}p{\dimexpr.82\linewidth/11-2\tabcolsep\relax}}@{\hspace{\tabcolsep}}}
\toprule
& \multicolumn{6}{c}{\textbf{PhyGenBench}} & \multicolumn{5}{c}{\textbf{VideoPhy-2 Hard}}\\
\cmidrule(lr){2-7}\cmidrule(lr){8-12}
\textbf{Method} & Chem. & Force & Heat & Light & Phys. & \textbf{Overall} & SA & PC & SA$\geq4$ & PC$\geq4$ & \textbf{Overall}\\
\midrule
Random selection & \metriccell{tablegood!0!tablemid}{55.56} & \metriccell{tablepoor!33!tablemid}{63.33} & \metriccell{tablegood!100!tablemid}{60.74} & \metriccell{tablepoor!33!tablemid}{69.78} & \metriccell{tablepoor!100!tablemid}{54.67} & \metriccell{tablepoor!33!tablemid}{62.78} & \metriccell{tablepoor!33!tablemid}{3.11} & \metriccell{tablepoor!100!tablemid}{4.23} & \metriccell{tablepoor!33!tablemid}{38.15} & \metriccell{tablepoor!100!tablemid}{77.59} & \metriccell{tablepoor!100!tablemid}{30.56}\\
WMReward~\cite{yuan2026inference} & \metriccell{tablegood!0!tablemid}{55.56} & \metriccell{tablepoor!100!tablemid}{62.50} & \metriccell{tablepoor!100!tablemid}{54.44} & \metriccell{tablepoor!100!tablemid}{68.67} & \metriccell{tablegood!0!tablemid}{56.00} & \metriccell{tablepoor!100!tablemid}{61.25} & \metriccell{tablepoor!100!tablemid}{3.06} & \metriccell{tablepoor!100!tablemid}{4.23} & \metriccell{tablepoor!100!tablemid}{37.78} & \metriccell{tablepoor!33!tablemid}{77.78} & \metriccell{tablegood!0!tablemid}{\underline{32.22}}\\
GeoPhys~\cite{interno2026geophys} & \metriccell{tablepoor!100!tablemid}{53.33} & \metriccell{tablegood!33!tablemid}{65.83} & \metriccell{tablegood!33!tablemid}{60.00} & \metriccell{tablegood!100!tablemid}{72.67} & \metriccell{tablegood!100!tablemid}{58.67} & \metriccell{tablegood!100!tablemid}{\textbf{64.58}} & \metriccell{tablegood!100!tablemid}{3.14} & \metriccell{tablegood!0!tablemid}{4.31} & \metriccell{tablegood!33!tablemid}{38.89} & \metriccell{tablegood!100!tablemid}{81.11} & \metriccell{tablegood!0!tablemid}{\underline{32.22}}\\
SPAR & \metriccell{tablegood!100!tablemid}{57.78} & \metriccell{tablegood!100!tablemid}{66.67} & \metriccell{tablepoor!33!tablemid}{57.78} & \metriccell{tablegood!33!tablemid}{71.33} & \metriccell{tablegood!100!tablemid}{58.67} & \metriccell{tablegood!33!tablemid}{\underline{64.38}} & \metriccell{tablegood!33!tablemid}{3.12} & \metriccell{tablegood!100!tablemid}{4.36} & \metriccell{tablegood!100!tablemid}{39.44} & \metriccell{tablegood!33!tablemid}{80.56} & \metriccell{tablegood!100!tablemid}{\textbf{33.89}}\\
\bottomrule
\end{tabular*}
\end{table}

Table~\ref{tab:supp-wan-transfer} shows that SPAR improves over both random selection and WMReward on the two T2V benchmarks with the unchanged reward configuration.
On PhyGenBench, it reaches 64.38, improving over WMReward by 3.13 points.
The improvement over WMReward extends across all five physical categories.
On VideoPhy-2 Hard, it achieves the highest joint success among the compared selectors, 33.89\%, compared with 32.22\% for both WMReward and GeoPhys.
SPAR also obtains the highest mean PC score and semantic-adherence pass rate among these selectors.
These results support transferring the self-referenced reward from MiniMax-H3 to Wan without adapting its coefficients.

\subsection{Candidate-Budget Scaling on Physics-IQ}
\label{sec:supp-budget-scaling}

We examine how a larger candidate pool affects physical alignment using Wan2.2-TI2V-5B on the 198 Physics-IQ inputs spanning 66 scenes.
The evaluated budgets are $N\in\{1,3,5,7\}$.
For each input, larger budgets extend the same ordered candidate pool, and every selector ranks the identical first $N$ videos.
SPAR retains the MiniMax-H3 reward configuration throughout.
Scores follow the official Physics-IQ Verified evaluation on a 0--100 scale.
Random selection reports the exact uniform expectation at each budget; at $N=1$, all methods return the same video.

\begin{table}[ht]
\centering
\caption{\textbf{Candidate-budget scaling on Physics-IQ Verified with Wan~\cite{wan2025wan}.} Scores (left) and their trends (right) use nested candidate pools with fixed reward parameters. For $N>1$, bold marks the highest score at each budget; all methods coincide at $N=1$.}
\label{tab:supp-budget-scaling}
\begin{minipage}[c]{0.56\linewidth}
\centering
\small
\renewcommand{\arraystretch}{1.15}
\setlength{\tabcolsep}{4pt}
\begin{tabular*}{0.92\linewidth}{@{\extracolsep{\fill}}lrrrr@{}}
\toprule
Selection rule & $N=1$ & $N=3$ & $N=5$ & $N=7$\\
\midrule
Random selection & 30.79 & 30.23 & 29.58 & 29.89\\
WMReward~\cite{yuan2026inference} & 30.79 & 31.66 & 31.75 & 33.48\\
GeoPhys~\cite{interno2026geophys} & 30.79 & \textbf{32.16} & 32.27 & 33.05\\
SPAR & 30.79 & 31.62 & \textbf{32.55} & \textbf{33.76}\\
\bottomrule
\end{tabular*}
\end{minipage}\hfill
\begin{minipage}[c]{0.40\linewidth}
\centering
\includegraphics[width=\linewidth]{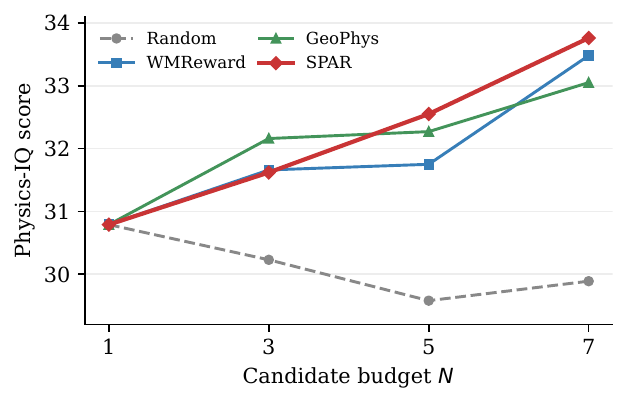}
\end{minipage}
\end{table}

SPAR's score increases across the evaluated budgets from 30.79 at $N=1$ to 33.76 at $N=7$ (Table~\ref{tab:supp-budget-scaling}).
At $N=5$ and $N=7$, SPAR obtains the highest scores among the compared selectors, while the random-selection expectation remains below the single-candidate score.
At $N=7$, SPAR exceeds random selection by 3.87 points and the single-candidate result by 2.97 points.
These results indicate that the fixed reward can use additional candidates to improve physical continuation quality.

\section{Component Analysis and Robustness}
\label{sec:full-ablation}

This section examines reference assignment (Section~\ref{sec:supp-referencecontrol}), individual reward components (Section~\ref{sec:supp-component-ablations}), and the role of the history-contrast sign (Section~\ref{sec:history-sign}).
It further examines the parameters governing spatial and temporal aggregation (Section~\ref{sec:supp-robustness}) and the coefficients combining the reward components, including their stability and retuning after component removal (Section~\ref{sec:fusion-search}).

\subsection{Reference Assignment}
\label{sec:supp-referencecontrol}
\label{sec:referencecontrol}

We test whether the benefit of static calibration depends on matching each reference to its own candidate.
The linear fusion in SPAR separates into a real-video contribution and a static-reference contribution.
Using the focused error $q_w$ defined in the main paper, these contributions are
\begin{align}
A_i={}&-\langle q\rangle_{\mathrm{r},i}
+\beta P(v_i)+\gamma\langle d\rangle_{\mathrm{r},i},\\
F_i={}&\langle q\rangle_{\mathrm{f},i}
-\gamma\langle d\rangle_{\mathrm{f},i}.
\end{align}
Thus $R(v_i)=A_i+F_i$, and $A_i$ is the reward without static calibration.

For the three candidates indexed by $i\in\{1,2,3\}$, the two permutations assigning every candidate another candidate's reference are $(\pi(1),\pi(2),\pi(3))=(2,3,1)$ and $(3,1,2)$.
Each permutation assigns scores $A_i+F_{\pi(i)}$ and selects the highest-scoring candidate.
We average the evaluation scores of the two resulting selections for each input, then aggregate across inputs.
Both static anchors are reassigned together, while the real-video contributions remain fixed.

The shared-reference control uses $A_i+\frac{1}{3}\sum_{j=1}^{3} F_j$ and selects exactly the same candidates as $A_i$ on all 538 inputs, as expected from invariance to a common offset.
The static-only control selects by $F_i$.
All controls retain the same candidates and reward coefficients, isolating the effect of reference assignment.

\begin{table}[t]
\centering
\caption{\textbf{Candidate-specific reference control.} Real-video contributions and coefficients are fixed. Exchanged references average the two within-input derangements. A shared reference gives the same selections as no reference. VideoPhy-2 Hard reports GPT-6 Sol joint success in percent.}
\label{tab:supp-referencecontrol}
\label{tab:referencecontrol}
\small
\setlength{\tabcolsep}{4pt}
\begin{tabular*}{\linewidth}{@{\extracolsep{\fill}}lrrr@{}}
\toprule
{\bfseries\boldmath Reference assignment} & {\bfseries\boldmath Physics-IQ Verified $\uparrow$} & {\bfseries\boldmath PhyGenBench $\uparrow$} & {\bfseries\boldmath VideoPhy-2 Hard (\%) $\uparrow$}\\
\midrule
Static contribution alone & 40.75 & \underline{63.75} & \textbf{62.78}\\
W/o candidate-specific reference & \underline{43.85} & 62.08 & 61.11\\
W/ exchanged references & 42.97 & 61.98 & \underline{61.94}\\
Full (SPAR) & \textbf{44.47} & \textbf{65.00} & \textbf{62.78}\\
\bottomrule
\end{tabular*}
\end{table}

Table~\ref{tab:supp-referencecontrol} shows that matching each reference to its own candidate improves over removing or exchanging references on all three benchmarks.
The improvement over exchanged references is 1.51 points on Physics-IQ, 3.02 points on PhyGenBench, and 0.83 percentage points on VideoPhy-2 Hard.
A shared reference leaves the candidate ranking unchanged. Selecting by the static contribution alone matches SPAR on VideoPhy-2 Hard but yields lower Physics-IQ and PhyGenBench scores.
Together, these controls support using the candidate's own baseline in conjunction with its observed dynamics.

\subsection{Component Ablations}
\label{sec:supp-component-ablations}

We extend the component analysis in Table~\mainpaperref{tab:ablation} to examine reference construction, spatial weighting, and temporal refinement under fixed coefficients.
Removing static references omits their contributions to both prediction-error scoring and history contrast; the single-anchor variants instead use only the first or middle frame throughout the reward.
Spatial refinement is removed by setting $\alpha=0$, while hard-only and continuous-only weighting correspond to $\lambda=0$ and $\lambda=1$.
Peak compensation and history contrast are removed by setting $\beta=0$ and $\gamma=0$, respectively; setting both to zero removes temporal refinement.
We also examine history contrast without static calibration, retaining the comparison between long- and short-history predictions only on the original video.
All other coefficients remain fixed at their full-method values.

\begin{table}[t]
\centering
\caption{\textbf{Fixed-parameter ablations.} W/o: without. W/: with the specified replacement. Each row changes the named component and keeps the remaining coefficients fixed. VideoPhy-2 Hard reports GPT-6 Sol joint success in percent.}
\label{tab:supp-ablation}
\small
\setlength{\tabcolsep}{4pt}
\begin{tabular*}{\linewidth}{@{\extracolsep{\fill}}lrrr@{}}
\toprule
{\bfseries\boldmath Configuration} & {\bfseries\boldmath Physics-IQ Verified $\uparrow$} & {\bfseries\boldmath PhyGenBench $\uparrow$} & {\bfseries\boldmath VideoPhy-2 Hard (\%) $\uparrow$}\\
\midrule
W/o static references & 43.85 & 62.08 & 61.11\\
W/ first anchor only & 43.24 & \underline{64.58} & 59.44\\
W/ middle anchor only & 43.18 & 64.17 & 61.67\\
W/o spatial refinement & 43.71 & \underline{64.58} & 60.56\\
W/ hard selection only & 44.22 & 63.96 & \textbf{62.78}\\
W/ continuous weights only & 43.88 & \underline{64.58} & \underline{62.22}\\
W/o temporal refinement & 43.81 & 62.29 & 61.11\\
W/o peak compensation & 43.78 & 63.33 & 61.11\\
W/o history contrast & 44.29 & 62.71 & \underline{62.22}\\
W/ uncalibrated history contrast & \underline{44.30} & 62.71 & \underline{62.22}\\
Full (SPAR) & \textbf{44.47} & \textbf{65.00} & \textbf{62.78}\\
\bottomrule
\end{tabular*}
\end{table}

\paragraph{Static calibration across reward components.}
SPAR-base, the basic self-referenced reward defined in the main paper, improves over uniform random selection on all three benchmarks.
The complete method further improves these results by incorporating spatial and temporal refinement (Table~\ref{tab:supp-progressive}).
Within the complete method, dual-anchor references outperform both the uncalibrated reward and either single-anchor alternative on every benchmark (Table~\ref{tab:supp-ablation}).
On Physics-IQ, the gains over the first-frame and middle-frame references are 1.24 and 1.29 points, respectively, supporting the use of complementary static anchors.

\paragraph{Complementary spatial weighting.}
Spatial refinement improves Physics-IQ by 0.76 points and VideoPhy-2 joint success by 2.22 percentage points relative to uniform spatial averaging.
Combining hard and continuous weighting yields higher Physics-IQ and PhyGenBench scores than either scheme alone and matches hard-only weighting on VideoPhy-2.
These comparisons support combining emphasis on the largest errors with graded weighting across spatial positions.

\paragraph{Complementary temporal evidence.}
Peak compensation and history contrast each improve all three benchmark scores when added to the other components (Table~\ref{tab:supp-ablation}).
History contrast without its static reference yields nearly the same results as omitting history contrast, supporting calibration of this term against the candidate's static response.
The complete reward also outperforms the partial combinations in Table~\ref{tab:supp-progressive}, supporting the joint use of spatial and temporal evidence.

\begin{table}[ht]
\centering
\caption{\textbf{Progressive combinations.} All coefficients retain their full-method values. Physics-IQ Verified and PhyGenBench use 0--100 scales; VideoPhy-2 Hard reports GPT-6 Sol joint success in percent.}
\label{tab:supp-progressive}
\small
\setlength{\tabcolsep}{4pt}
\begin{tabular*}{\linewidth}{@{\extracolsep{\fill}}lrrr@{}}
\toprule
{\bfseries\boldmath Configuration} & {\bfseries\boldmath Physics-IQ Verified $\uparrow$} & {\bfseries\boldmath PhyGenBench $\uparrow$} & {\bfseries\boldmath VideoPhy-2 Hard (\%) $\uparrow$}\\
\midrule
SPAR-base & 43.49 & 64.17 & 60.56\\
SPAR-base + Spatial & 43.81 & 62.29 & 61.11\\
SPAR-base + Spatial + Peak compensation & \underline{44.29} & 62.71 & \underline{62.22}\\
SPAR-base + Temporal & 43.71 & \underline{64.58} & 60.56\\
Full (SPAR) & \textbf{44.47} & \textbf{65.00} & \textbf{62.78}\\
\bottomrule
\end{tabular*}
\end{table}

\subsection{History-Contrast Sign and Static Response}
\label{sec:history-sign}
The sign of $\gamma$ determines how history contrast contributes to candidate ranking.
We compare the default negative coefficient with zero and an equal-magnitude positive coefficient, holding all other coefficients fixed.
For $\gamma=-0.06875$, $0$, and $+0.06875$, respectively, Physics-IQ scores are $44.47$, $44.29$, and $44.05$, and PhyGenEval open scores are $65.00$, $62.71$, and $62.50$.
The corresponding VideoPhy-2 GPT-6 Sol joint-success rates are $62.78\%$, $62.22\%$, and $62.22\%$.
The default negative coefficient gives the highest scores on all three benchmarks.
Across the 1,614 candidates, the mean difference in agreement between long and short histories is $0.1485$ on static references and $0.0133$ on real windows.
Thus, additional history can improve feature-change agreement even for static inputs.
With $D(v)=\langle d\rangle_{\mathrm r}-\langle d\rangle_{\mathrm f}$, a negative coefficient favors a smaller increase in agreement on the original video relative to its static references.
The sign therefore governs the contribution of relative context sensitivity to the reward, rather than expressing a preference for a particular history length.

\subsection{Aggregation-Parameter Sensitivity}
\label{sec:supp-robustness}

The aggregation parameters determine how predictive evidence is summarized within each reward component.
We vary the number of high-error spatial positions $K$, the continuous-weight cap $c$, and the number of peak windows $m$ individually, holding the fusion coefficients $(\alpha,\lambda,\beta,\gamma)$ at their default values (Table~\ref{tab:supp-sensitivity}).
Continuous-weight caps of two, three, four, and eight yield Physics-IQ scores between 44.38 and 44.47 and VideoPhy-2 joint-success rates between 62.22\% and 62.78\%.
The default configuration achieves the highest Physics-IQ score and matches the highest PhyGenBench and VideoPhy-2 scores among the tested settings.
\begin{table}[ht]
\centering
\caption{\textbf{Single-factor sensitivity.} One structural setting changes at a time; the four fusion coefficients are fixed. The default uses 64 tail positions, weight cap four, and two peak windows. VideoPhy-2 Hard reports GPT-6 Sol joint success in percent.}
\label{tab:supp-sensitivity}
\small
\setlength{\tabcolsep}{4pt}
\begin{tabular*}{\linewidth}{@{\extracolsep{\fill}}lrrr@{}}
\toprule
{\bfseries\boldmath Setting} & {\bfseries\boldmath Physics-IQ Verified $\uparrow$} & {\bfseries\boldmath PhyGenBench $\uparrow$} & {\bfseries\boldmath VideoPhy-2 Hard (\%) $\uparrow$}\\
\midrule
Tail size 32 & 43.58 & \underline{64.79} & 61.67\\
Tail size 96 & 44.00 & 64.38 & 61.67\\
Tail size 128 & 43.73 & \textbf{65.00} & 61.11\\
Weight cap 2 & 44.38 & \underline{64.79} & \underline{62.22}\\
Weight cap 3 & \underline{44.42} & \underline{64.79} & \textbf{62.78}\\
Weight cap 8 & \underline{44.42} & \textbf{65.00} & \textbf{62.78}\\
One peak window & 43.98 & 63.75 & 60.56\\
Three peak windows & 44.37 & 63.33 & 61.67\\
Five peak windows & 43.78 & 63.33 & 61.11\\
Full (SPAR) & \textbf{44.47} & \textbf{65.00} & \textbf{62.78}\\
\bottomrule
\end{tabular*}
\end{table}

At the boundary $m=W$, $\operatorname{TopMean}_W(E)=\langle E\rangle_{\mathrm r}$ and hence $P(v)=0$, making this setting equivalent to removing peak compensation.
With $\beta$ fixed, the single-peak case extends compensation to a signed mean-error weighting, whereas the default two-peak case retains nonnegative weights.

\subsection{Fusion-Coefficient Analysis}
\label{sec:fusion-search}
The fusion coefficients determine the relative contributions of the reward components.
With the aggregation parameters fixed at $K=64$, $c=4$, and $m=2$, we examine candidate-selection stability as $(\alpha,\lambda,\beta,\gamma)$ vary and whether retuning the remaining coefficients can recover the benefits of a removed component.

\paragraph{Coefficient grid and optimization protocol.}
The grid uses the following values:
\begin{align*}
\alpha\in{}&\{0,.2,.35,.5,.6,.675,.75,.85,1,1.25\},\\
\beta\in{}&\{0,.25,.4,.5,.6,.7,.85,1,1.25\},\\
\gamma\in{}&\{-.2,-.15,-.1,-.075,-.06875,\\
&\hspace{5mm}-.05,-.025,0,.025,.05\},\\
\lambda\in{}&\{0,.125,.25,.375,.5,.625,.75,.875,1\}.
\end{align*}
The search includes spatial extrapolation ($\alpha>1$) and signed temporal weighting ($\beta>m/(W-m)$), extending beyond the nonnegative mixtures used by the default configuration.
The full-method grid contains 8,100 combinations; each ablation uses the same grid values for its active coefficients.
A local grid further evaluates $5^4=625$ settings around the default full-method configuration.
For $(\alpha,\beta,\gamma,\lambda)$, the step sizes are $(0.00625,0.00625,0.0015625,0.03125)$, with offsets of zero, one, or two steps in either direction.

\paragraph{Candidate-selection stability.}
Fifteen distinct coefficient settings select exactly the same candidates for all 538 inputs, matching the default configuration $(\alpha,\lambda,\beta,\gamma)=(.675,.25,.60,-.06875)$.
Their marginal parameter ranges are $\alpha\in[.6625,.6875]$, $\beta\in[.60,.60625]$, $\gamma\in[-.0703125,-.065625]$, and $\lambda\in[.1875,.3125]$.
These are marginal ranges over the 15 settings, not a continuous region in which every parameter combination produces identical selections.

\paragraph{Ablations with retuned coefficients.}
To assess whether coefficient adjustment can compensate for removing a component, we reoptimize the remaining fusion coefficients for each ablation.
The configurations follow the component definitions in Table~\mainpaperref{tab:ablation}; removing spatial refinement sets $\alpha=0$, making $\lambda$ inactive.
Table~\ref{tab:supp-reoptimized} reports the resulting cross-benchmark tradeoffs.
Retuning reveals tradeoffs across benchmarks: hard-only weighting reaches 44.62 on Physics-IQ, compared with 44.47 for SPAR, while attaining 62.71 on PhyGenBench and 61.11\% on VideoPhy-2, compared with 65.00 and 62.78\% for SPAR.
Across the evaluated grids, no ablated configuration matches or exceeds the full method on all three benchmarks simultaneously, supporting the combined reward beyond a single fixed coefficient setting.

\begin{table*}[t]
\centering
\caption{\textbf{Ablations with retuned coefficients.} Each ablation reports the configuration with the highest Physics-IQ score in the common grid over its active coefficients. Full (SPAR) uses the default coefficients; the additional local grid applies only to the full method. VideoPhy-2 Hard reports GPT-6 Sol joint success in percent.}
\label{tab:supp-reoptimized}
\small
\setlength{\tabcolsep}{4pt}
\begin{tabular*}{\linewidth}{@{\extracolsep{\fill}}lrrrrrrr@{}}
\toprule
{\bfseries\boldmath Configuration} & {\bfseries\boldmath $\alpha$} & {\bfseries\boldmath $\beta$} & {\bfseries\boldmath $\gamma$} & {\bfseries\boldmath $\lambda$} & {\bfseries\boldmath Physics-IQ Verified $\uparrow$} & {\bfseries\boldmath PhyGenBench $\uparrow$} & {\bfseries\boldmath VideoPhy-2 Hard (\%) $\uparrow$}\\
\midrule
W/o static references & .50 & .60 & -.10 & 1 & 43.94 & 62.50 & \underline{62.22}\\
First-frame reference only & .50 & .85 & .025 & .125 & 44.14 & \underline{64.58} & 60.56\\
Middle-frame reference only & .20 & .60 & .05 & .125 & 43.84 & 62.50 & 61.67\\
W/o spatial refinement & 0 & .50 & -.025 & --- & 43.86 & \underline{64.58} & 60.56\\
Hard-only weighting & .675 & .50 & -.025 & 0 & \textbf{44.62} & 62.71 & 61.11\\
Continuous-only weighting & .85 & .60 & -.025 & 1 & 44.28 & 63.96 & \underline{62.22}\\
W/o peak compensation & 1 & 0 & .025 & 0 & 44.16 & 62.71 & \underline{62.22}\\
W/o history contrast & .75 & .50 & 0 & 0 & 44.45 & 62.71 & 61.67\\
Full (SPAR) & .675 & .60 & -.06875 & .25 & \underline{44.47} & \textbf{65.00} & \textbf{62.78}\\
\bottomrule
\end{tabular*}
\end{table*}

\clearpage
\begin{figure}[p]
\centering
\includegraphics[width=\textwidth,height=0.9\textheight,keepaspectratio]{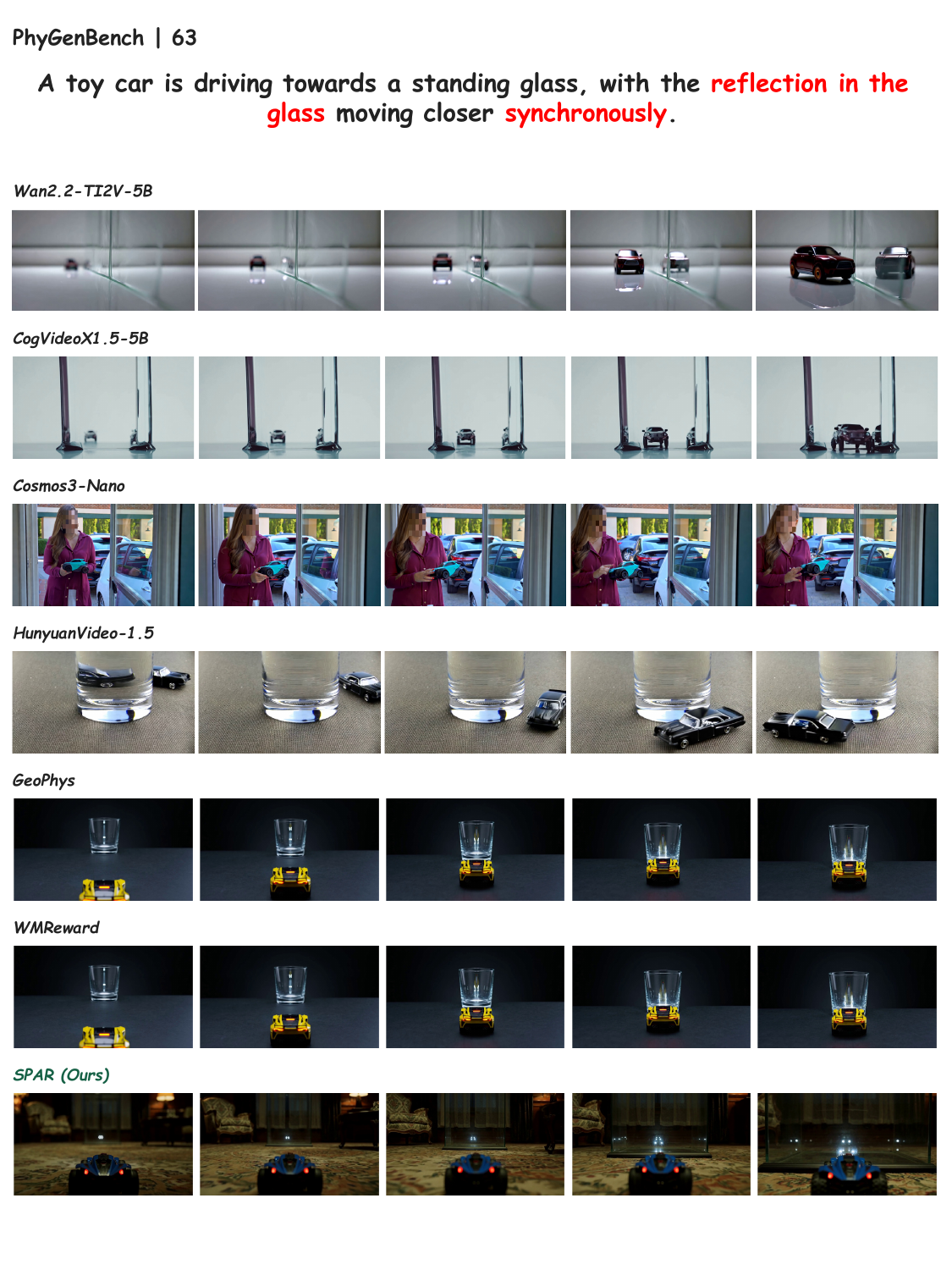}
\caption{\textbf{Qualitative comparison on PhyGenBench: reflection during motion.} The prompt describes a toy car approaching a glass, with its reflection moving synchronously.}
\label{fig:supp-qualitative-1}
\end{figure}

\clearpage
\begin{figure}[p]
\centering
\includegraphics[width=\textwidth,height=0.9\textheight,keepaspectratio]{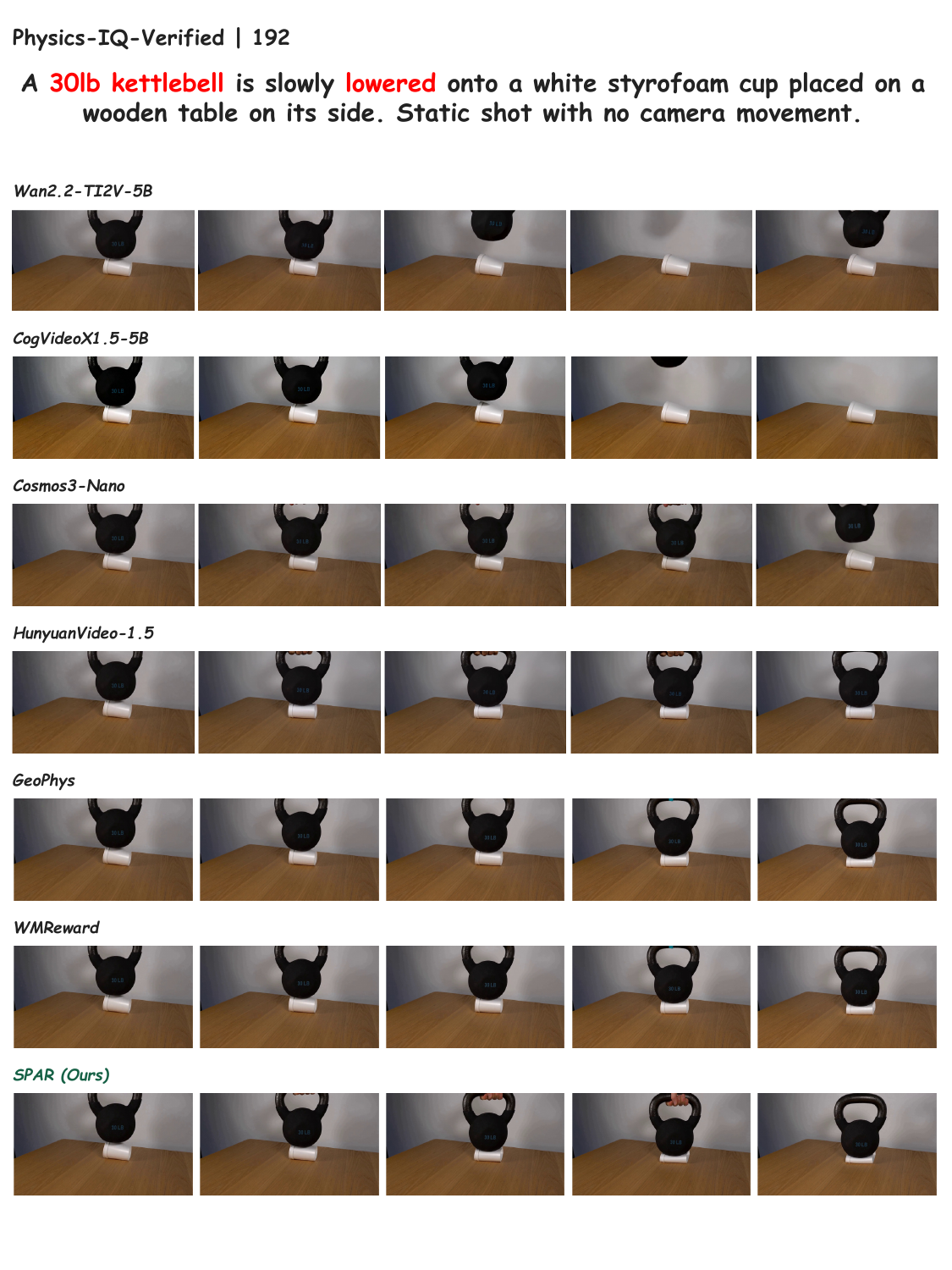}
\caption{\textbf{Qualitative comparison on Physics-IQ Verified: compression under load.} The prompt describes a kettlebell being lowered onto a styrofoam cup.}
\label{fig:supp-qualitative-2}
\end{figure}

\clearpage
\begin{figure}[p]
\centering
\includegraphics[width=\textwidth,height=0.9\textheight,keepaspectratio]{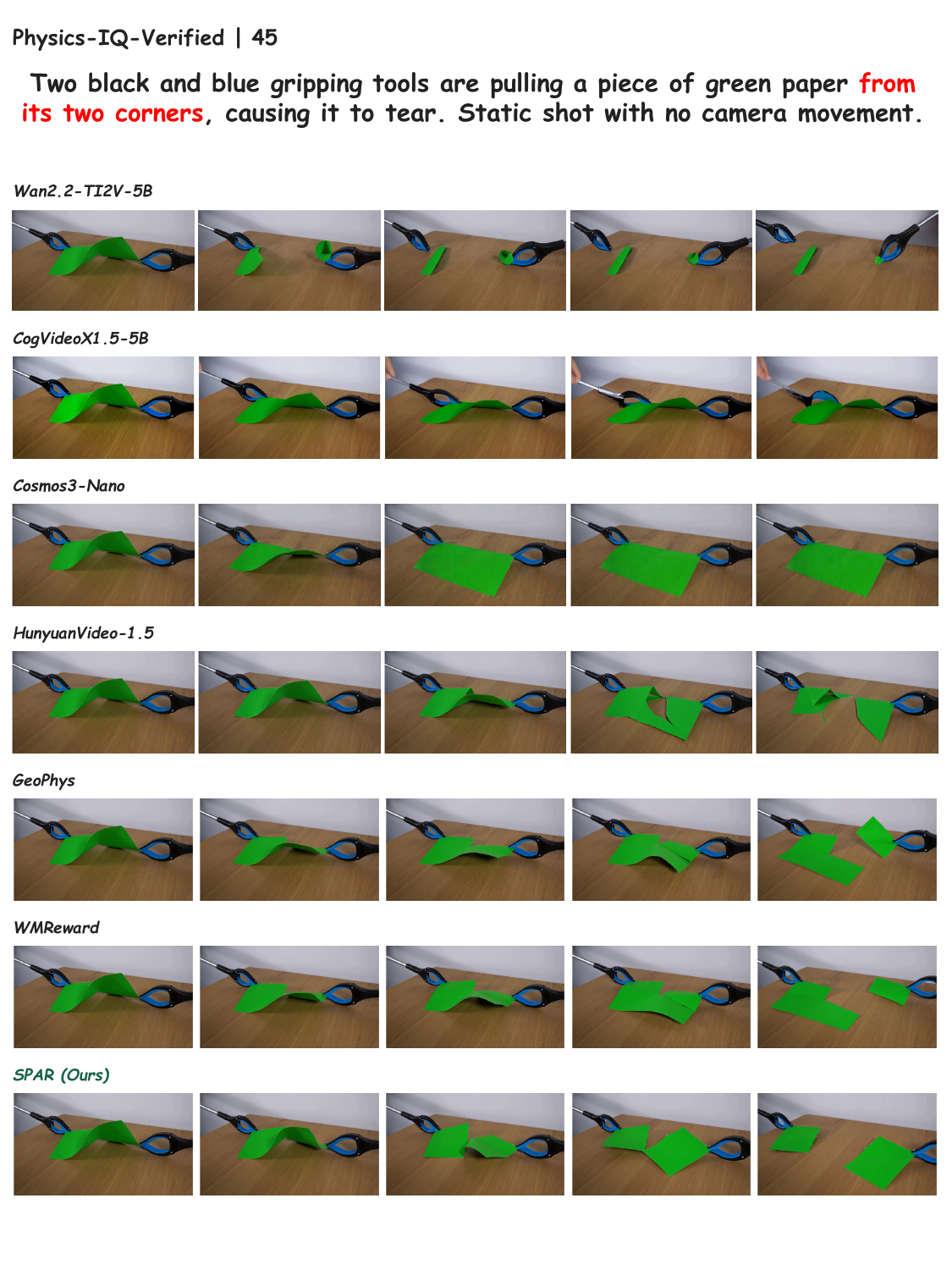}
\caption{\textbf{Qualitative comparison on Physics-IQ Verified: tearing under tension.} The prompt describes two gripping tools pulling a sheet of paper apart.}
\label{fig:supp-qualitative-3}
\end{figure}

\clearpage
\begin{figure}[p]
\centering
\includegraphics[width=\textwidth,height=0.9\textheight,keepaspectratio]{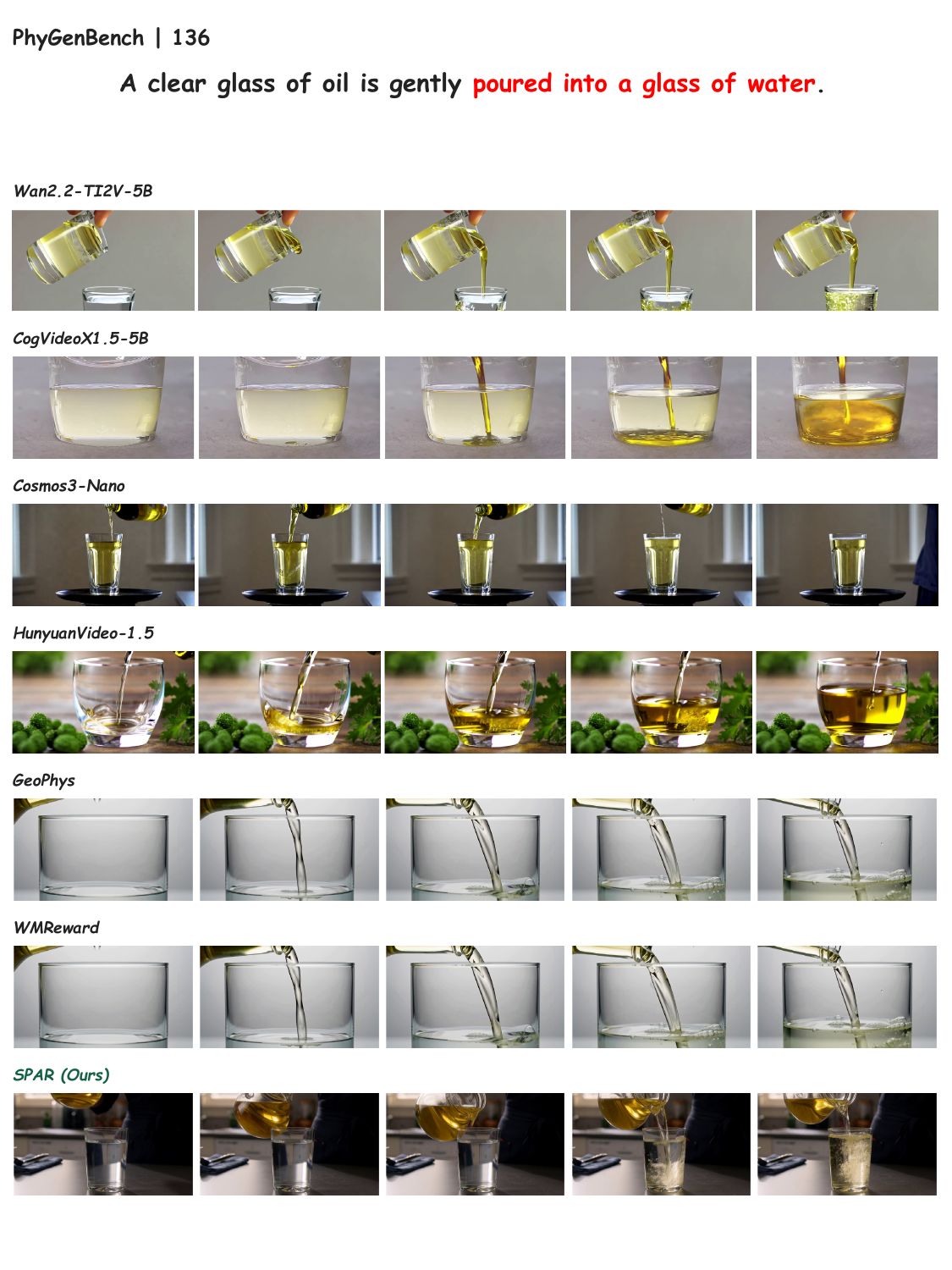}
\caption{\textbf{Qualitative comparison on PhyGenBench: pouring oil into water.} The prompt describes oil being gently poured into a glass of water.}
\label{fig:supp-qualitative-4}
\end{figure}

\clearpage
\begin{figure}[p]
\centering
\includegraphics[width=\textwidth,height=0.9\textheight,keepaspectratio]{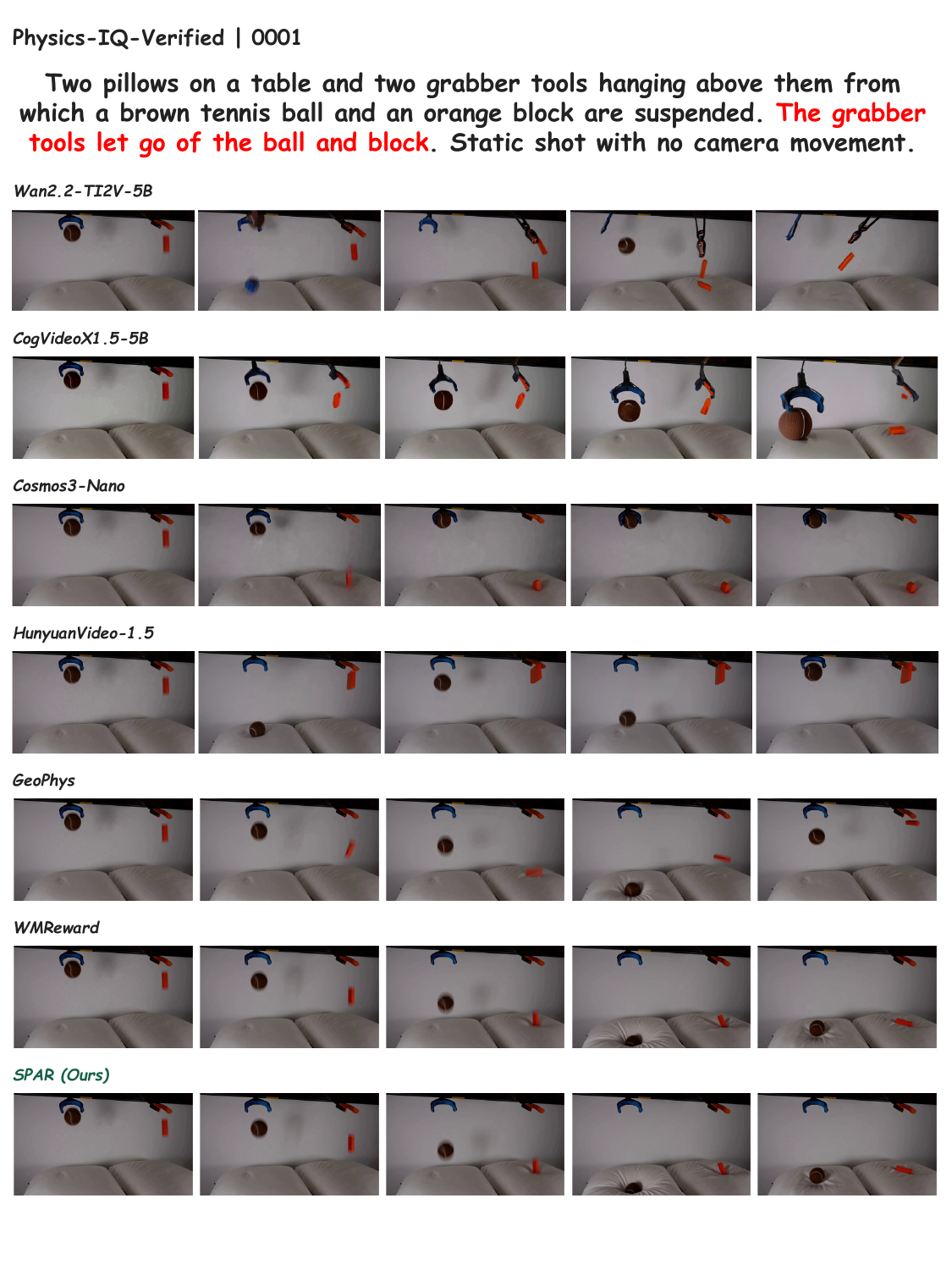}
\caption{\textbf{Qualitative comparison on Physics-IQ Verified: object release.} The prompt describes a tennis ball and a block being released above two pillows.}
\label{fig:supp-qualitative-5}
\end{figure}

\clearpage
\begin{figure}[p]
\centering
\includegraphics[width=\textwidth,height=0.9\textheight,keepaspectratio]{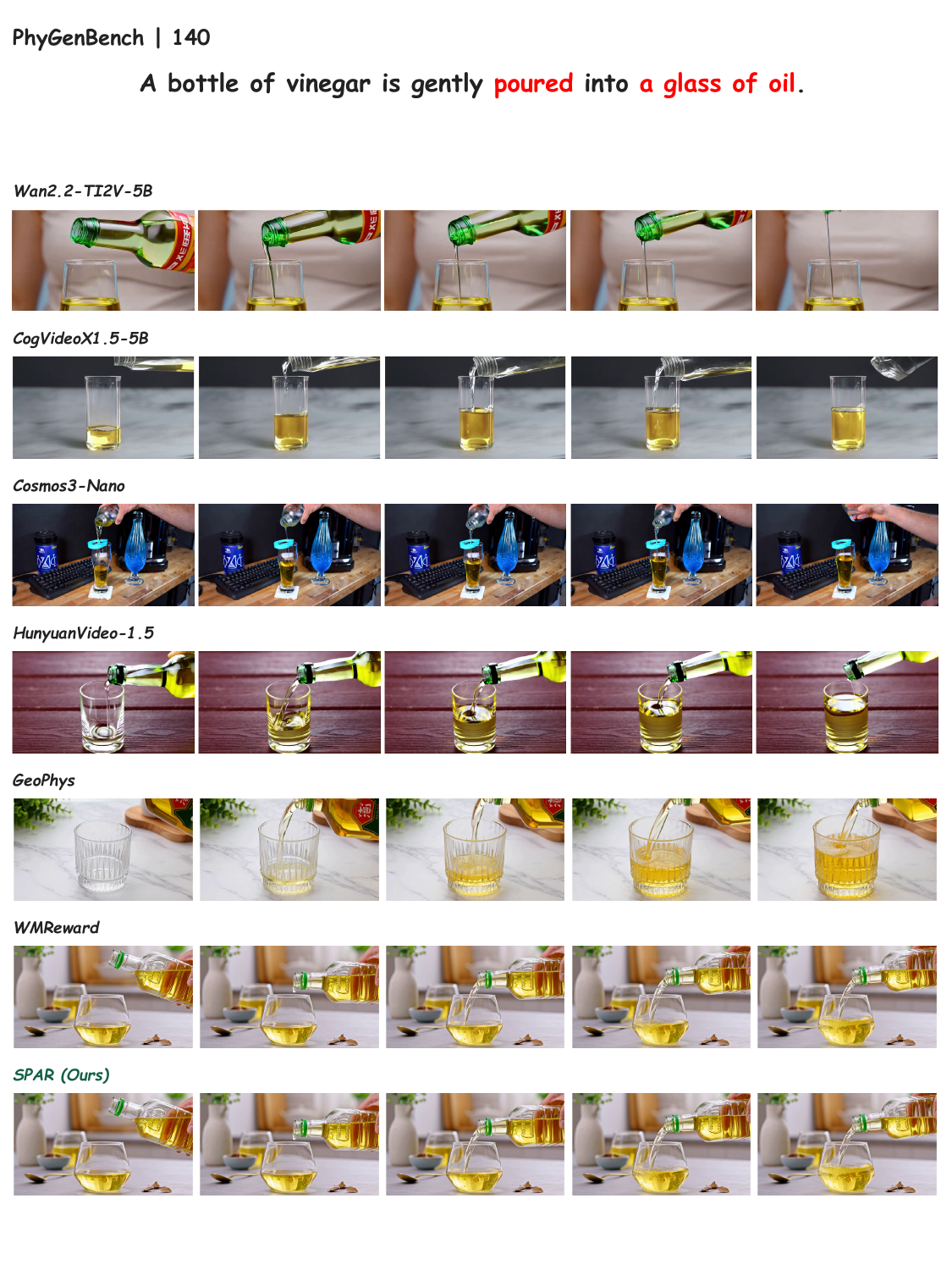}
\caption{\textbf{Qualitative comparison on PhyGenBench: pouring vinegar into oil.} The prompt describes vinegar being gently poured into a glass of oil.}
\label{fig:supp-qualitative-6}
\end{figure}

\clearpage
\begin{figure}[p]
\centering
\includegraphics[width=\textwidth,height=0.9\textheight,keepaspectratio]{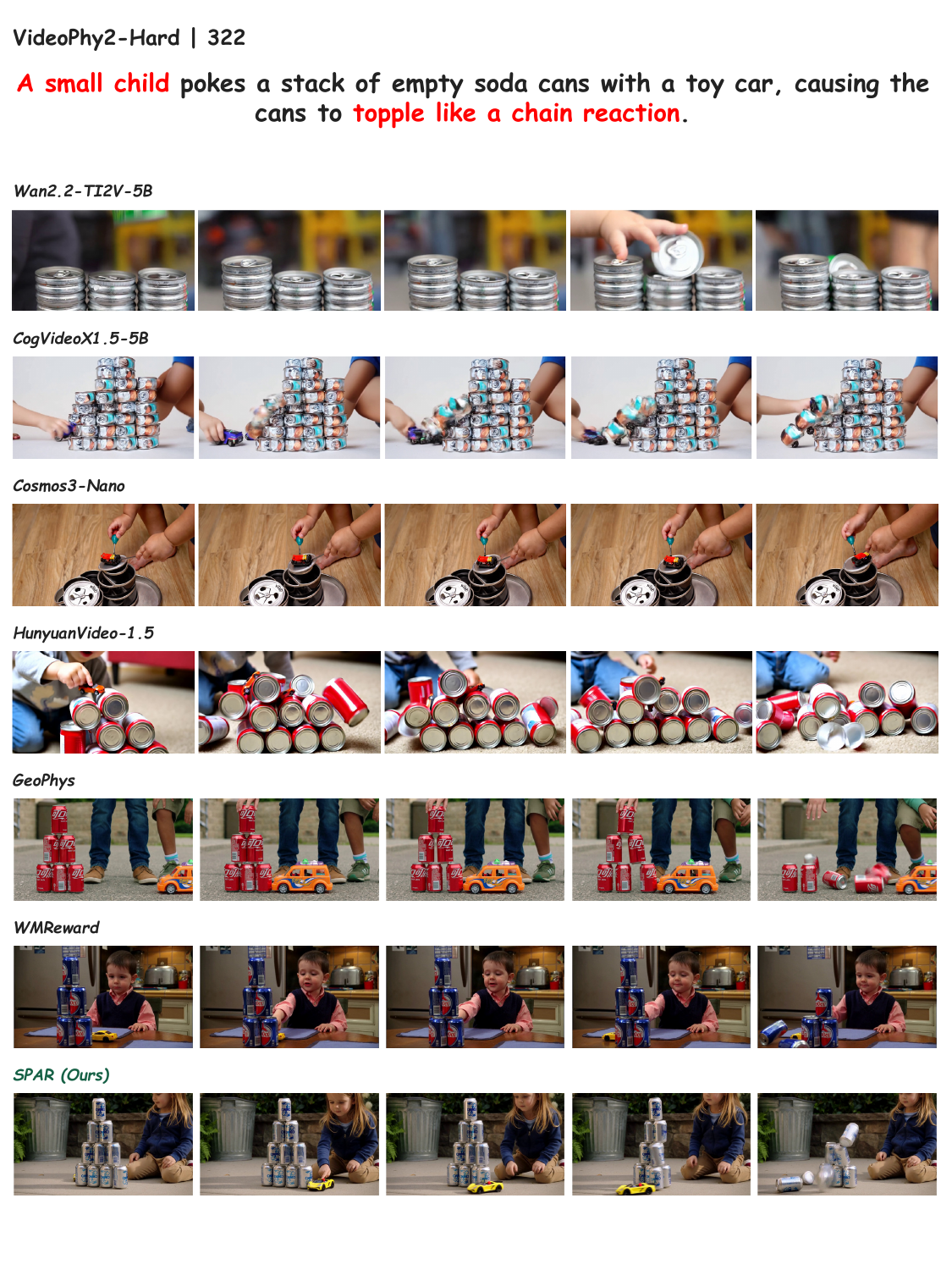}
\caption{\textbf{Qualitative comparison on VideoPhy-2 Hard: sequential toppling.} The prompt describes a child pushing a stack of empty cans with a toy car, causing a chain reaction.}
\label{fig:supp-qualitative-7}
\end{figure}
\clearpage

\end{document}